%% file: aaai25.tex
\documentclass[letterpaper]{article} % DO NOT CHANGE THIS
\usepackage{aaai25}  % DO NOT CHANGE THIS
\usepackage{times}  % DO NOT CHANGE THIS
\usepackage{helvet}  % DO NOT CHANGE THIS
\usepackage{courier}  % DO NOT CHANGE THIS
\usepackage[hyphens]{url}  % DO NOT CHANGE THIS
\usepackage{graphicx} % DO NOT CHANGE THIS
\usepackage{natbib}  % DO NOT CHANGE THIS AND DO NOT ADD ANY OPTIONS TO IT
\usepackage{caption} % DO NOT CHANGE THIS AND DO NOT ADD ANY OPTIONS TO IT
\usepackage{amsmath}
\usepackage{tabularray}
\usepackage{algorithm}
\usepackage{algorithmic}
\usepackage{multirow}
\usepackage{graphicx}  % For \rotatebox
\usepackage{array}     % For table enhancements
\usepackage{amssymb}
\usepackage{tikz-dependency}
\usepackage{pgfplots}
\usepackage{pgfplotstable}
\usepackage{subfig}
\usepackage{pifont}       % \ding{xx}
\usepackage{bbding}
\pgfplotsset{compat=newest}

\usepackage{xcolor}

\usepackage{newfloat}
\usepackage{listings}
\DeclareCaptionStyle{ruled}{labelfont=normalfont,labelsep=colon,strut=off} % DO NOT CHANGE THIS
\floatstyle{ruled}
\newfloat{listing}{tb}{lst}{}
\floatname{listing}{Listing}
\title{Dynamic Adapter with Semantics Disentangling \\ for Cross-lingual
Cross-modal Retrieval}

\author{
    Rui Cai\textsuperscript{\rm 1,2},
    Zhiyu Dong\textsuperscript{\rm 1,2},
    Jianfeng Dong\textsuperscript{\rm 1,2}\thanks{The corresponding author.},
    Xun Wang\textsuperscript{\rm 1,2}\\
}
\affiliations{
    \textsuperscript{\rm 1}  the College of Computer Science and Technology, Zhejiang Gongshang University, Hangzhou, China\\
    \textsuperscript{\rm 2} Zhejiang Key Laboratory of Big Data and Future E-Commerce Technology,
     Hangzhou, China\\
     cairuics@gmail.com, dongzhiyuuu@163.com, \{djf, wx\}@zjgsu.edu.cn
}

\usepackage{bibentry}
\begin{document}

\maketitle

\begin{abstract}

Existing cross-modal retrieval methods typically rely on large-scale vision-language pair data. This makes it challenging to efficiently develop a cross-modal retrieval model for under-resourced languages of interest.
Therefore, Cross-lingual Cross-modal Retrieval~(CCR), which aims to align vision and the low-resource language (the \textit{target language}) without using any human-labeled target-language data, has gained increasing attention. 
As a general parameter-efficient way, a common solution is to utilize adapter modules to transfer the vision-language alignment ability of Vision-Language Pretraining (VLP) models from a source language to a target language.
However, these adapters are usually \textit{static} once learned,  making it difficult to adapt to target-language captions with \textit{varied} expressions.
To alleviate it, we propose \textit{\textbf{D}ynamic \textbf{A}dapter with \textbf{S}emantics \textbf{D}isentangling}~(DASD), 
whose parameters are dynamically generated conditioned on the characteristics of the input captions.
Considering that the semantics and expression styles of the input caption largely influence how to encode it, we propose a semantic disentangling module to extract the semantic-related and semantic-agnostic features from the input, ensuring that generated adapters are well-suited to the characteristics of input caption.
Extensive experiments on two image-text datasets and one video-text dataset demonstrate the effectiveness of our model for cross-lingual cross-modal retrieval, as well as its good compatibility with various VLP models.
%Our source code is anonymously released at \url{https://github.com/HuiGuanLab/DASD}.

\end{abstract}

\begin{links}
    \link{Code}{https://github.com/HuiGuanLab/DASD}
\end{links}

\input{intro}
\input{relwork}
\input{method}
\input{eval}
\input{conlusion}
\input{acknowledgement}
\bibliography{aaai25}
\clearpage
\begin{appendix}
\input{supplementary}
\end{appendix}

\end{document}

%% file: intro.tex
\section{Introduction}

%\tiny  \scriptsize \footnotesize \small
% \captionsetup[subfloat]{font=small} 
% \begin{figure}[tb!]
% \centering
% \vspace{-0mm}
% \subfloat[Captions with the same semantics but different expressions.]
% {\includegraphics[width=0.45\textwidth]{images/image_forms.png}}%
% \hfil
% \vspace{-3mm}
% \subfloat[A typical adapter for target-langauge encoding.]
% {\includegraphics[width=0.45\textwidth]{images/typical adapter.png}}%
% \hfil
% \vspace{-3mm}
% \subfloat[Our proposed DASD framework for CCR]
% {\includegraphics[width=0.45\textwidth]{images/DASD.png}}%
% \vspace{-3mm}
% \caption{Illustration of the variety of textual expressions and the difference between the traditional adapter and our DASD: (a) Captions of the same image are differently expressed in Chinese-specific ways.
% (b) Traditional adapters whose parameters are static during the inference. (c) Our proposed DASD  extracts semantic-related (SR) and semantic-agnostic (SA) features from captions and thereby produces dynamic adapters (DA).}
% \vspace{-4mm}
% \label{fig:intro_figure}
% \end{figure}

\begin{figure}[tb!]
\centering
\includegraphics[width=1\columnwidth]{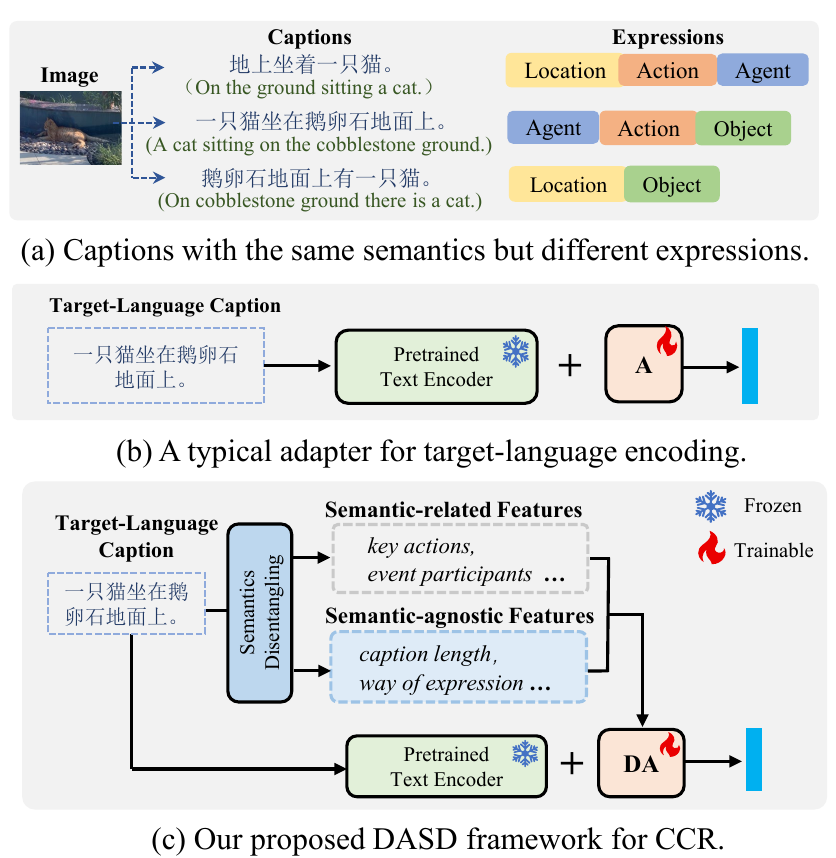}
%\vspace{-4mm}
\caption{Illustration of the variety of textual expressions and the difference between the traditional adapter and our DASD: (a) Captions of the same image are differently expressed in Chinese-specific ways.
(b) Traditional adapters whose parameters are fixed once learned. (c) Our method extracts semantic-related and semantic-agnostic features from captions and thereby produces dynamic adapters (DA).} 
%\vspace{-4mm}
\label{fig:intro_figure}
\end{figure}

With the rapid emergence of images and videos on the Internet, there is a huge demand from users around the world for retrieving visual content of interest by natural language queries (\textit{a.k.a.} cross-modal retrieval)~\cite{li2021adaptive, zhang2022modality,chang2023learning}.
Recent neural-based cross-modal retrieval models~\cite{bogolin2022cross,lu2022cots,sun2023hierarchical} tend to require a large amount of human-labeled text-image/video pair data for training which are available for only a handful of the world’s languages.
As a result, building a cross-modal retrieval system for users with different language backgrounds is extremely challenging, especially for low-resource languages (\textit{e.g.,} Czech). 
%With this regard, \textit{Cross-lingual} Cross-modal Retrieval (CCR), especially the one
%transferring the advantage of the source language with affluent amount of resources (\textit{e.g.,} English)
%to the target language where the labeled data is scarce or even not available, is of great importance.
%To build cross-modal retrieval models for other languages, \textit{cross-lingual} cross-modal retrieval~(CCR), especially the one transferring from the source language with affluent amount of resources~(\textit{e.g.,} English) to the target language where the human-labeled data is scarce or even not available, is of great importance.
With this regard, \textit{cross-lingual} cross-modal retrieval~(CCR) leverages visual-text pair data in the rich-resource language (the \textit{source language}) to construct a retrieval model for a new language of interest (the \textit{target language}), 
%without using any human-labeled target-language data.
avoiding substantial manual annotations costs on the target language. 

The perennial problem with building target-language retrieval models lies in the paucity of training data,
since existing human-labeled resources for low-resource languages are rather limited, and it is extremely expensive and time-consuming to manually annotate images/videos with descriptions in multiple languages.
%To alleviate this problem, a series of Multilingual Vision-Language Pre-training (M-VLP) models \cite{ni2021m3p,zhou2021uc2, zeng2022cclm}, which handle multiple languages and modalities simultaneously during pre-training, have been proposed and achieved convincing performance on cross-lingual image and video retrieval tasks.
Due to limitations in data and computing resources, existing Vision-Language Pretraining (VLP) models, such as CLIP~\cite{radford2021learning} and CCLM~\cite{zeng2023cclm}, could support only one or a few languages, yet there are more than 6,900 languages worldwide~\cite{zhou2021uc2}.
Moreover, these VLP models cannot be flexibly extended to new languages, since additional training on target languages will cause performance degeneration of VLP models on the original languages due to the limited model capacity.

A straightforward and cheap solution is converting source-language labeled data into the target language utilizing Machine Translation (MT) tools (\textit{e.g.,} Google Translate\footnote{\url{https://translate.google.com/}}). 
With access to these MT-generated resources, existing works tend to transfer the vision-language alignment ability of VLP models to target languages through cross-lingual alignment~\cite{wang2024cl2cm, wang2024dual,pfeiffer2020mad,zhang2022multi}.
Among them, a prior work~\cite{wang2022cross} tries to finetune the pre-trained layers in VLP models with cross-lingual alignment objectives,
inevitably leading to a certain degree of knowledge forgetting. 
To alleviate this problem, some adapter-based methods~\cite{pfeiffer2020mad, zhang2022multi} %and prompt learning~\cite{yang2024m3p} 
have recently been proposed to perform cross-lingual transfer in a parameter-efficient way. %, extending VLP models to under-represented languages.
These methods freeze VLP models 
and store cross-lingual knowledge in the light-weight adapters, whose
parameters keep static for different inputs.
However, during the cross-lingual transfer, the language gaps~\cite{ahmad-etal-2019-cross}, such as unique expressions in target languages,
%and word ordering distances~\cite{ahmad-etal-2019-cross}, 
could bring complexity and increase the difficulty of extracting the accurate semantics of captions.
As illustrated in Figure~\ref{fig:intro_figure}(a),  target-language (Chinese) captions describing the same event are expressed in quite different ways.
%some of which might have never appeared during the source-language pretraining owning to the language gaps.
As a result, existing \textit{static} adapters, shown in Figure~\ref{fig:intro_figure}(b), struggle to adapt to target-language captions with \textit{varied} expressions. % and unseen expressions, 
%since it could be hardly achieved with rather limited trainable parameters which have no relevance to the inputs.
%since their trainable parameters are static and rather limited to keep the parameter efficiency of the entire model.
%As a result, such challenges demand a more dynamic and adaptive tuning approach.
%Some recent efforts~\cite{wang2022cross,wang2023cl2cm,wang2024dual} tries to alleviate this problem by automatically filtering translation errors brought by MT during the interaction between source-language and  target-language sentences.
%Nevertheless, these approaches only focus on reducing the impact of obvious errors brought by MT, neglecting other factors (e.g., ambiguous sentences) which could also hurt the cross-lingual alignment and result in degraded performance.
%Some  means based on adapters~\cite{zhang2022multi} and prompt learning~\cite{yang2024m3p} has recently been proposed to perform cross-lingual transfer in a parameter-efficient way, however, they all fail to reduce the impact of uncertainties, since it could be hardly achieved with limited trainable parameters which are fixed during the inference.

To tackle the aforementioned challenge,  we propose \textit{Dynamic Adapter with Semantics Disentangling} (DASD), a novel paradigm that adaptively encodes target-language captions by making language adapters conditioned on each input caption rather than keeping them fixed after once learning.
%to dynamically learn the cross-lingual correspondences with an ideal trade-off between task performance and parameter efficiency.
%In order to adaptively encoding captions with different semantics and expressions, 
%To achieve this, DASD employs dynamic parameters which are contextually generated according to the characteristics of input captions.
%The key idea is to make adapters conditioned on the characteristics of each input instance (caption) rather than fixed once learned.
In order to obtain adapters that exactly match the input
caption, we perform semantics disentangling to capture its distinct and complementary aspects.
To be specific, we assume that each caption is entangled by two independent characteristics: semantic-related and semantic-agnostic features.
Particularly, the former presents consistent semantic features shared by different modalities, while the latter reflects the characteristics with respect to the mean of expression yet unrelated to semantics,
such as word order, sentence length, and other low-level information (shown in Figure~\ref{fig:intro_figure}(c)).
%Although the essence of CCR is extracting accurate and consistent semantic representations from data in different modalities, we argue that semantic-unrelated features also play an important role in characterizing a caption.
%In this paper, the semantics disentanglement is achieved by performing adversarial training, and both semantic-related and semantic-unrelated features are captured in a parameter-efficient way.
%by the first $K$ layers of the frozen VLP model equipped with light-weight adapters.
%Surprisingly, we found that even a small value of $K$ ($K$ is 1 in our model) can still achieve a comparable performance to the using full model, achieving an ideal trade-off between task performance and parameter efficiency.
Both semantic-related and semantic-agnostic features of input captions are learned by semantics disentangling, which explicitly decouples the two different features through semantic consistency learning and adversarial training.
In this way, the disentangled features capture sufficient information needed for characterizing the input caption, which are then fed to the dynamic parameter generation module.
Cross-lingual alignments are finally performed with dynamic parameters inserted to adapters, thus allowing the model to encode captions while explicitly accounting for the overall characteristics observed in the textual inputs.

%automatically fixing translation errors in MT-generated sentences is a non-trivial task which relies on the supervision from a certain amount of original-translated sentences pairs, and such resources are unfortunately not available for low-resource languages.
%Besides, 

To the best of our knowledge, this is the first work that leverages disentangled semantics to generate dynamic adapters to improve the target-language text encoding in CCR.
Our main contributions are summarized as follows:
\begin{itemize}
\item We identify the problem of performing accurate text encoding against challenges
caused by the diversity in written expression  of target-language training samples in CCR, 
and provide an effective solution based on data-dependent semantics disentanglement.

\item We propose a novel parameter-efficient diagram for CCR which dynamically generates parameters of input-aware adapters, enabling the CCR models to encode target-language sentences adaptively.

\item We achieve a new state-of-the-art performance on two image-text retrieval datasets and one video-text retrieval dataset. Besides, our model shows good compatibility with various VLP models.
% We conduct extensive experiments on multiple benchmarks, where our proposed model achieves a new state-of-the-art (SOTA). 
% Experimental results also demonstrate that our proposed method is compatible with different VLP models.
\end{itemize}
%dynamically generate data-dependent adapters and thereby extract more accurate semantics from the textual inputs. Our main contributions are summarized as follows:

%% file: relwork.tex
\section{Related Work}

\subsection{Cross-lingual Cross-modal Retrieval}
Cross-lingual cross-modal retrieval (CCR) is a method for achieving visual and target language (V-T) alignment without using any manually-annotated visual-text data pairs.
This approach can be seen as a specific case of transfer learning to new domains~\cite{fang2024your} with limited resources~\cite{zhang2020causal},  helping to mitigate the scarcity of training data for low-resource languages in traditional cross-modal retrieval~\cite{dong2022partially, dong2022dual,fang2023you,zheng2023progressive}.
Early works~\cite{aggarwal2020towards,portaz2019image} on CCR try to transfer the knowledge of English models to low-resource languages by directly finetuning the model on MT-generated parallel data. 
Recently, V+L pretraining models have become popular, aiming to further narrow the gap between different languages and modalities.
Among them, M$^3$P~\cite{ni2021m3p} learns universal representations that can map objects occurred in different modalities or texts expressed in different languages into a common semantic space. UC$^2$~\cite{zhou2021uc2} translates source-language annotations into the target language automatically and proposes fine-grained pretraining objectives to encourage alignment between image regions and multilingual tokens.
Following UC$^2$, MURAL~\cite{jain2021mural} leverages 1.8 billion noisy image-text pairs to pre-train their dual encoder model. 
After that, CCLM~\cite{zeng2023cclm} proposes a cross-view language modeling framework, which considers both multi-modal data and multi-lingual data as pairs of two different views of the same object and propose a unified framework to fuse features in different views. Although CCLM successfully outperforms UC2 and MURAL on several benchmarks, it is very expensive to expand CCLM to support new low-resource languages since its pretraining stage require large-amount data and computing resources. %considering both multi-modal data and multi-lingual data as two different views of the same object.
%, and trains the model to align the two views by maximizing the mutual information between them with conditional Masked Language Modeling (MLM) and contrastive learning.
Instead of pretraining V+L models from scratch, some recent works~\cite{wang2022cross,wang2024cl2cm,wang2024dual, cai2024cross} try to finetune the upper layers of existing VLP models with machine-translated data, which inevitably leading to a certain degree of knowledge forgetting.

Although these works have achieved improvements on CCR, their methods still require full-model training, which is quite time-consuming and demands significant computational power, making them impractical for researchers with limited hardware resources.

\subsection{Parameter-Efficient FineTuning for CCR}
The pretraining and finetuning paradigms have been proven to be highly effective in different language and vision tasks. Compared to full fine-tuning, Parameter-Efficient FineTuning (PEFT) is more suitable for cases with limited hardware resources, as it freezes the majority of the parameters of the pretrained model while still being able to demonstrate comparable performance in downstream tasks. 
Various PEFT techniques have been explored, including prompt tuning~\cite{li2021prefix,liu2023gpt,wu2024mitigate,zhou2022conditional}, Low-Rank Adaptation (LoRA)~\cite{hedegaard2024structured,mao2024dora}, and adapters~\cite{zhang2022multi}.
In which, the core idea of adapters is to insert light-weight adaptation modules into each layer of the pretrained transformer~\cite{vaswani2017attention}, and they have been extended across numerous domains. For example, MAD-X~\cite{pfeiffer2020mad} extends multilingual pretraining models to support low-resource languages through adapters. 
Following MAD-X, MAD-G~\cite{ansell2021mad} is proposed to generate language adapters based on type characteristics in language representations.

%However, these methods are built upon VLPs trained
%relies on MLM-based VLPs and cannot be applied to dual-stream models such as CLIP, they are not is not suitable for cross-modal retrieval considering the gap between retrieval and MLM.

Recently, MLA~\cite{zhang2022multi} designs a light-weight language acquisition encoder that supports low-resource languages through language-specific adapters.
This approach is somewhat similar to our idea but has some core differences: MLA overlook the diversity of written expression and the noise in translated training samples during the cross-lingual transfer. 
In contrast, our DASD learns disentangled characteristics of input captions to help the model understand sentences in different styles of expression.
% The issue of how to utilize semantically unrelated information has already been thoroughly studied in the field of natural language processing. In the past, it was frequently used in tasks involving generation. In the field of audio, it is interpreted as information on the acoustic features~\cite{yan2024talk}.
% This work argues that the same sentence can convey different meanings when spoken with different intonations, thus interpreting semantic-unrelate representations as speaking styles. 
% In addition, when generating text~\cite{chanthran2024bridging}, semantic-unrelate information is also used to improve the grammatical correctness of the generated text.
% In the task of Font Generation, MX-Font \cite{park2021multiple} is trained to learn by separating the content and style from the training data.

% In CCR tasks, to our knowledge, no articles have utilized semantic-unrelate information. Most practices leverage language-agnostic representation  to facilitate cross-language transfer~\cite{wang2022cross}. Our work attempts to fill this gap in research by using adversarial learning during the pre-reading phase to let the adapter learn semantic-unrelated representations from the sentence text, and attempts to use this semantic-unrelate information to help the model understand the textual sentences.
\subsection{Semantics Disentangling}
Recently, learning disentangled representations has been widely applied to  a wide spectrum of applications ranging from domain adaption~\cite{cai2019learning, zhang2024disentangling} to text-to-image generation~\cite{yin2019semantics} and zero-shot learning~\cite{chen2021semantics,ye2021disentangling}.
The core idea behind these work is to factorize input features into semantic-related and semantic-unrelated representations, so that the disentangled semantic-related features could be adapted across domains, modalities or tasks. 
For example, the prior work~\cite{yin2019semantics} focusing on text-to-image generation distills semantic commons from the linguistic descriptions, based on which the generated images can keep
generation consistency under expression variants.
Different from these previous approaches, in this paper, semantic-unrelated representations also play an important role during the transfer across languages, which are utilized for the dynamic adapter generation to improve the semantics extraction of target-language captions.

%% file: method.tex
\begin{figure*}[tb!]\centering\includegraphics[width=1.95\columnwidth]{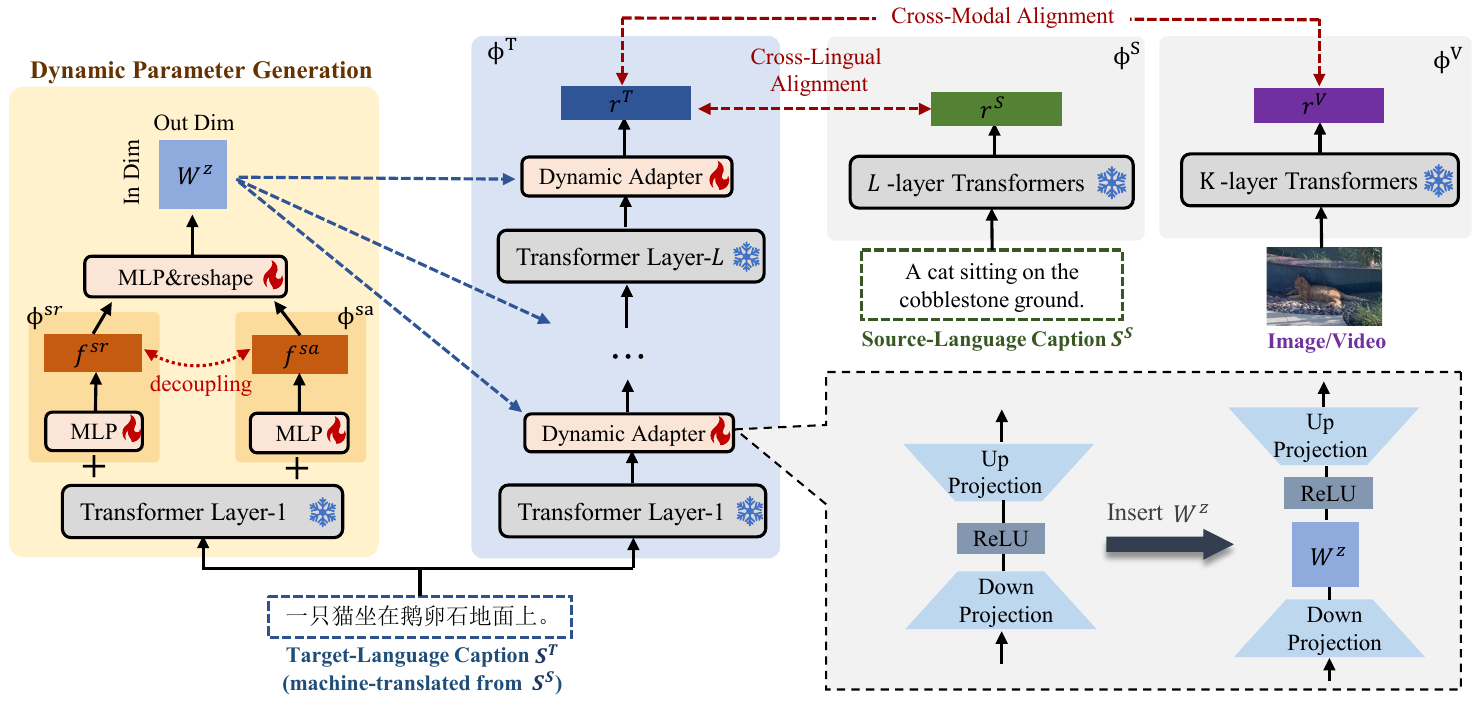}
%\vspace{-1mm}
\caption{The illustration of our proposed Dynamic Adapter with Semantics Disentangling (DASD).
To make dynamic adapters in the target-language branch $\Phi^T$ exactly match its input $S^T$, semantics disentangling is performed to extract semantic-related and semantic-agnostic features ($f^{sr}$ and $f^{sa}$) from $S^T$ and then generate input-conditional parameters (shown in the leftmost branch).
The source-language branch $\Phi^S$ and visual branch $\Phi^V$ are provided by the frozen VLP model. 
%The leftmost branch extract semantic-related and semantic-agnostic features from the target-language input $S^T$ and generates dynamic parameter matrix $W^z$ conditioned on the input, which are inserted to the input-aware dynamic adapters of the target-language branch $\Phi^T$.}\label{fig:framework}
%\vspace{-5mm}
}
\label{fig:framework}
\end{figure*}

\section{The Proposed Method}
%The dynamic adapter with semantics disentangling (DASD) is proposed to empower a
%existing VLP model with cross-lingual ability by adaptively generating parameters of adapters
%considering the characteristics of the text inputs. 
%Following MLA~\cite{zhang2022multi}, we choose CLIP~\cite{radford2021learning} as the VLP model, and our method can also be applied to other VLP models with different native languages.
%In what follows, we first introduce the task definition and a basic model for CCR, followed by the description of semantic distribution estimation module.
%Then, in Section, we illustrate how to generate parameters of context-aware adapters
%according to the estimated semantics.
%Finally, we present the overall training and inference strategy. 
In this paper, we propose a dynamic adapter generation framework with semantics disentangling for CCR.
As shown in Figure~\ref{fig:framework}, our framework consists of three key components:
1) a pretrained VLP model as the backbone of our framework whose parameters stay frozen;
2) an input-aware parameter generator which analyzes the characteristics of the target-language input and produces a parameter matrix of adapters accordingly;
3) dynamic adapters inserted to each layer of the frozen VLP model to adaptively empower it with the cross-lingual ability.

\subsection{Task Definition}
We first formally define the setting of CCR, which involves two kinds of languages, namely the source language and the target languages.
%which involves two languages, namely the source language $S$ and the target language $T$.
For the source language $S$, we have a collection of human-labeled
training data $\mathcal{D}^S = \{d_1,d_2, ... , d_n\}$, where each instance $d_i$ consists of a caption $S^S_i$ paired with an image or video $V_i$.
As for the target language $T$, due to the scarcity of human-labeled
data, we assume there are no extra labeled data of text-image/video pairs.
The core task of CCR is to obtain a model applicable in the target language $T$, without using any manually annotated target-language data.

\subsection{Pretrained VLP Model}
Following MLA~\cite{zhang2022multi}, we choose CLIP~\cite{radford2021learning} as the VLP model used in our DASD. 
It is worth noting that other VLP models can also be applied to our method.
%with different native languages.

%An additional [cls] token is added before the image patches, and its output at the last Transformer layer represents the image’s global feature.
\subsubsection{Source-language Text Encoding.} Given a sentence $S^S$ in the source language, the corresponding sentence
representation $r^S = \Phi^S(S^S; \theta^S)$ is generated through the pretrained text encoder $\Phi^S$, which contains a  embedding block and $L$ transformer layers. 
To preserve the cross-model knowledge of VLP, $\theta^S$ keeps fixed during training.
Concretely, the input sentence $S^S$ is tokenized and processed into word embeddings 
$E^S = [e_{0=\mathtt{[SOS]}},...,e_{M=\mathtt{[EOS]}}]$ through the embedding block, where $\mathtt{[SOS]}$ and $\mathtt{[EOS]}$ are special tokens denoting the boundary 
of the input sentence.
The word embeddings are then fed to the parameter-frozen CLIP's text encoder.  %equipped with the distribution adapter $\theta^{\mu}$:
%\begin{align}
%    X_0^{S}&=[e_{0=\mathtt{[SOS]}}, e_1, ..., e_{M=\mathtt{[EOS]}}] + E_{pos}\\
%    X_i^{S}&=\mathtt{TransformerLayer}(X_{i-1}^{S}) 
    %X_i^{S} &=  \mathtt{DA}(H_i^{S};\theta^{S}_i)
%\end{align}
%where $X^S_i = [x^S_{i,0}, ...,x^S_{i,M}]$ is the hidden state of the $i$-th layer, and $E_{pos}$ is the positional encoding.
The final representation $r^S$ is obtained by performing a linear projection on the last hidden state of the $\mathtt{[EOS]}$ token.
%: $r^S = W^a x^S_{l,M}$.

\subsubsection{Visual Encoding.} 
The Vision Transformer (ViT)~\cite{dosovitskiy2020image, zhang2020feature} is used as a kind of CLIP image encoder, which takes image patches as input and generates the final feature through a Transformer-based model. 
For image encoding, given an image $V$,
it is divided into patches $V^{'} = [v^{'}_1, ..., v^{'}_N]$ following ViT. 
Then, they are linearly projected into patch embeddings $E_p=[e_{\mathtt{[CLASS]}}, W_p v^{'}_1, ... , W_p v^{'}_N]$, where $e_{\mathtt{[CLASS]}}$ is a
special embedding for the whole image and $W_p$ is the linear projection. 
The hidden states calculation is similar with the text encoder, and the final visual representation  $r^{V}$ is obtained by performing a linear projection on the last hidden state of the  $e_{\mathtt{[CLASS]}}$ token: $r^{V} = W^b h^l_0$.
As for video encoding, following the prior work~\cite{luo2022clip4clip}, we uniformly sample 12 frames from the video and then perform average pooling over these frame representations to obtain the final video representation.
%The image encoder of the pretrained
%CLIP~\cite{radford2021learning}  has aligned with the English text encoder by contrastive learning over 400M English image-text pairs, and it keeps frozen in our model.
%Φ denotes the parameters of the layer i.
%The CLIP text encoder has a similar structure to the GPT (Radford et al., 2019) model, whose final output of the [eos] token represents the global feature of an English sentence.
%Note that CLIP text encoder is directly used for English sentence encoding in our framework, and the parameters of both CLIP’s text encoder and image encoder are consistently frozen during the training and inference.

\subsection{Input-aware Dynamic Adapters}
%Given a target-language caption $S^T$, it is non-trivial to  extract its overall features with limited computational cost.
%In this paper, we propose to encode $S^T$ with only the first layer of $\Phi^S$ equipped with trainable adapters.
%In order to encode distinct and complementary aspects of $S^T$, we propose semantics disentangling to extract semantic-agnostic features  for caption encoding.
%, which could be extracted with the aid of adversarial training.
Given a target-language caption $S^T$ as the input, shown in the leftmost branch in Figure~\ref{fig:framework}, 
our method generates an input-conditional parameter matrix, which plays a key role in $S^T$ encoding.
%our method dynamically generates parameters according to the characteristics of $S^T$, enabling our model to encode $S^T$ adaptively. 
To generate parameters exactly match $S^T$,
we propose semantics disentangling to extract semantic-related and semantic-agnostic features from $S^T$.
\subsubsection{Semantics Disentangling.}
In the case of cross-lingual transfer, semantic-agnostic characteristics (such as the word order and the way of expression) in different languages tend to vary greatly, which can hardly be captured by static adapters. As a result, in our framework, we employ a semantics disentangling module to obtain semantic-agnostic features of captions in different target languages. 
Semantics disentangling performs feature extraction in a parameter-efficient way, whose backbone is only the first frozen layer of $\Phi^S$ equipped with two trainable lightweight adapters.
Taking $S^T$ as the input, semantic-related and semantic-agnostic features are extracted through semantic consistency learning and adversarial training, respectively.

\textit{\textbf{Semantic-Related Features Extraction.}}
% The essence of cross-modal retrieval is to capture the semantic consistency of paired heterogeneous data in the common space.
As shown in Figure~\ref{fig:framework}, the extraction of semantic-related features is performed by the module $\Phi^{sr}$, whose input is the target-language sentence $S^T$ which has been tokenized and processed into word embeddings 
$E^T = [u_{0=\mathtt{[SOS]}},...,u_{M=\mathtt{[EOS]}}]$. 
%We model the possible semantics of each target-language sentence $S^T$ with a Gaussian
%distribution $\mathcal N (\mu, \sigma^2)$, and the module to estimate the semantic distribution consists of two branches $\Phi_{\mu}$ and $\Phi_{\sigma}$ responsible for the mean $\mu$ estimation and the variance $\sigma$ estimation, respectively.
These embeddings are then fed to the first layer of parameter-frozen CLIP's text encoder equipped with
the semantic-related adapter $\mathtt{A^{sr}}$:
\begin{align}
    X_0^{sr}&=[W^{sr}_e u_{0}, W^{sr}_e u_1, ..., W^{sr}_e u_{M}] + E_{pos}\\
    H_1^{sr}&=\mathtt{TransformerLayer}(X_{0}^{sr}) \\
    X_1^{sr} &=  \mathtt{A^{sr}}(H_1^{sr};\theta^{sr}_1) + H^{sr}_1 \label{eq:Asr}
\end{align}
where $X_1^{sr}$ is the hidden state of the pretrained transformer layer and $W_e^{sr}$ is a linear projection to keep dimension consistency with source-language embeddings.
The semantic-related adapter $\mathtt{A^{sr}}$ in Equation~\ref{eq:Asr} is implemented as a bottleneck MLP with residual connection:
\begin{align}
 \mathtt{A^{sr}}(X) = W_{upper}^{sr} \mathtt{ReLU} (W_{down}^{sr} X)
\end{align}
Similar with $\Phi^S$, the last hidden state of the $\mathtt{[EOS]}$ token in the pretrained layer is linearly projected into the semantic-related feature $f^{sr}$:
\begin{align}
f^{sr}= W_{p}^{sr} x^{sr}_{1,\mathtt{[EOS]}}
\end{align}
Considering the fact that $S^T$ shares the same semantics with its source-language counterpart $S^S$, we propose semantic consistency learning to explicitly transfer the semantic information gathered by $\Phi^S$ from $S^S$ to $f^{sr}$.
%the  loss $\mathcal{L}_{kd}$ to strengthen the semantics decoupling,
%which introduces implicit knowledge in the sentence embedding $r^S$ to the extraction of semantic-related features.
As shown in Equation ~\ref{eq:l1},  the semantic consistency loss $\mathcal{L}_{sc}$ is defined as  the $L1$ distance between  $r^S$ and $ f^{sr}$:
\begin{equation}
\label{eq:l1}
    \begin{array}{cc}\
         \mathcal{L}_{sc} = ||r^S - f^{sr}||
    \end{array}
\end{equation}

\textit{\textbf{Semantic-Agnostic Features Extraction.}}
%Although semantic-agnostic information lack direct relations with semantic learning, we argue that semantic-agnostics features are also useful for encoding target-language sentences.
%play an equally important role as semantic-related features for text encoding in CCR task.
%Since the diversity of target-language expression poses challenges to adaptively encoding $S^T$ with only parameter-efficient modules, it is necessary to capture such semantically agnostic information with affordable computational  budgets.
%As shown in Figure~\ref{fig:framework}, 
The semantic-agnostic module $\Phi^{sa}$ shares the same backbone with $\Phi^{sr}$, equipped with the semantic-agnostic adapter $\mathtt{A^{sa}}$ which works in the similar way with $\mathtt{A^{sr}}$.
Different from $\Phi^{sr}$, $\Phi^{sa}$ produces the semantic-agnostic feature $f^{sa}$ by performing average-pooling over all hidden states in the pretrained transformer layer. 

To achieve perfect semantics disentangling, %adversarial training is facilitated to guarantee the feature $f^{sa}$ extracted from $S^T$ is semantic agnostic.
the semantic-agnostic feature $f^{sa}$ extracted from $S^T$ should exclude any semantic information about $S^T$. 
Motivated by this, we enforce $f^{sa}$ to be useless to identify its corresponding semantic representations through adversarial training.
Specifically, for adversarial training, we construct two feature pairs:($f^{sa}$,$r^S$) and ($f^{sa}$,$r^{S-}$).
The former is regarded as the positive sample since $r^S$ is the semantic representation of $S^S$ which shares the same semantics with $S^T$.
The latter is the negative sample and $r^{S-}$ is the semantic representation of a randomly-selected source-language caption.
%Specifically, given a source-language caption $S^S_i$ from $\mathcal{D^S}$ and a representation $r^{sa}_j$ generated by $\Phi^{sa}$, 
We employ a classifier $F$ to act as the discriminator which is adopted to distinguish the positive and negative samples. 
The classifier is consisted of a multi-layer feed-forward neural networks, and the discrimination loss in adversarial training is defined as:
\begin{equation}
    \mathcal{L}_{d} =  -log F(f^{sa}, r^S) - log(1 - F(f^{sa}, r^{S-}) \\
\end{equation}
The parameters of $\Phi^{sa}$ are updated to confuse the discriminator by minimizing the loss $\mathcal{L}_{adv} = -\mathcal{L}_d$.
%:
%\begin{align}
%r^{sa}= \mathtt{AvgPool} [x^{sa}_{1,0},x^{sa}_{1,1},...,x^{sa}_{1,M}]
%\end{align}
%We observe that the average pooling operation helps aggregating semantic-agnostic information more effectively than directly performing linear projection on $x^{sa}_{1,\mathtt{[EOS]}}$, leading to an improved performance.
%Another branch $\Phi^{\sigma}$ is also parameter-efficient implemented with the semantic acquirer $\mathtt{SA^{\sigma}}$.
%To ensure numerically stable training, instead of directly predicting $\sigma$,
%$\Phi_{\sigma}$ outputs the variance estimation: $s = log\sigma^2$ in the same way with $\Phi_{\mu}$.
%Then $\sigma$ is computed via an exponential mapping: $\sigma = exp(s/2)$, whose value always
%stays positive.

\subsubsection{Input-conditional Parameter Generation.}
After obtaining $f^{sr}$ and $f^{sa}$,  we adopt a multilayer perceptron to extract the global information $z$ of $S^T$, i.e.,
\begin{align}
z &= MLP(f^{sr} \circ f^{sa}) 
\end{align}
where $\circ$ means the concatenation operation.
Then, the dynamic parameter-matrix $W_i^{z}$ of layer $i$
are obtained  using a single layer linear down-projection:
\begin{align}
W_i^{z} &= reshape(W_i^{down}z) 
\end{align}
where $W_i^{down} \in \mathbb{R}^{{d^2_u} \times {d_z}}$, $W_i^{z} \in \mathbb{R}^{{d_u} \times {d_u}}$, and the operation $reshape$ refers to reshaping the vectors produced by the $MLP$ into a matrix form.
Down projecting to a dimension $d_u << d_z$ prevents $W_i^{down}$ from being impractically large, keeping our model parameter-efficient.
In this way, $W^z$ is dynamically generated conditioned on the target-language input $S^T$, which are then inserted to adapters in the target-language branch
$\Phi^T$.

\subsection{CCR with Dynamic Adapters}
The core of CCR is to align the target-language sentence $S^T$ with its source-language counterpart $S^S$ as well as its visual counterpart $V$.
However, due to the scarcity of human-labeled $S^T$-$V$ pairs, MT tools are employed to translate $S^S$ into the target language.
With access to these paired data from different sources, cross-lingual and cross-modal alignments are performed accordingly.

\subsubsection{Target-Language Text Encoding.}
As shown in Figure~\ref{fig:framework}, the representation of $S^T$ is calculated by the target-language branch $\Phi^T$, which taking word embeddings $E^T$ as the input and extract semantics of  $S^T$ at each layer with the help of the dynamic adapter $\mathtt{DA}$:
\begin{align}
    X^T_0&=[W^T_e u_{0=\mathtt{[SOS]}}, ..., W^T_e u_{M=\mathtt{[EOS]}}] + E_{pos}\\
    H^T_i&=\mathtt{TransformerLayer}(X^T_{i-1}) \\
    X^T_i &=   \mathtt{DA}(H^T_i;\theta^{\mathtt{DA}}_i) + H^T_i
\label{eq:lada}
\end{align}
where $\theta^{\mathtt{DA}}_i$ refers to the parameter of $\mathtt{DA}$ in $i$-th layer and works as follows:
\begin{align}
 \mathtt{DA}(X) = W_{upper}^{d} \mathtt{ReLU} (W^z W_{down}^{d} X)+X
\end{align}
Here, $\{W_{upper}^{d}, W_{down}^{d}, W^z\}\in \theta^{\mathtt{DA}}$.
Finally, the last hidden state of the $\mathtt{[EOS]}$ token is linearly projected into the semantic representation
of $S^T$: $r^T = W_p x^T_{\mathtt{[EOS]}}$.
The linear projection  $W_p$ is shared with CLIP's text encoder and keeps frozen during the training.

\subsubsection{Training Strategy.}
Considering the the scarcity of target-language resources, following MLA~\cite{zhang2022multi}, the cross-lingual alignment and the cross-modal alignment
are performed independently in our framework.
The motivation behind the separate training is to ensure that cross-lingual transfer can always proceed smoothly in case data in a certain modality is missing or of poor quality.
The objective in the cross-lingual alignment is minimizing the Mean Square Error
(MSE) between the native representation $r^S$ and the non-native representation $r^T$:
\begin{equation}
\label{eq:nlt}
    \begin{array}{cc}\
         \mathcal{L}_{CL} = ||r^S - r^T||^2
    \end{array}
\end{equation}
As for the cross-modal alignment, it is achieved by performing contrastive learning between target languages and images.
The training objective is minimizing the NCE loss~\cite{gutmann2010noise} defined as follows:
\begin{equation}
\label{eq:le}
\begin{split}
\mathcal{L}_{CM} = &- log\frac{{exp(
sim(r^{T},r^{V}) )}}{{\sum\nolimits_{j = 1}^B {exp(sim(r^{T}_j,r^{V}))} }} \\
 &- \log\frac{{exp(
sim(r^{T},r^{V}) )}}{{\sum\nolimits_{j = 1}^B {exp(sim(r^{T},r^{V}_j))} }}
\end{split}
\end{equation}
where $B$ is the batch size,  $sim(\cdot)$ denotes the similarity function (i.e., cosine similarity) and $\tau$ is the temperature coefficient.
Our model is trained by minimizing the combination of the 
above losses. Finally, the total loss function is defined as:
\begin{equation}
    \begin{array}{cc}
       \mathcal{L} =  \mathcal{L}_{CL} + \mathcal{L}_{CM} + \lambda_1 \mathcal{L}_{adv} +\lambda_2 \mathcal{L}_{sc}
      \label{eq:all}
    \end{array}
\end{equation}
where $\lambda_1$ and $\lambda_2$ are hyper-parameters to balance the importance of disentangling losses.

%% file: eval.tex
\section{Experiments}

\subsection{Experimental Settings}
\subsubsection{Datasets.}
Evaluations are performed on two image-text retrieval datasets~(Multi30K~\cite{elliott2016multi30k} and MSCOCO~\cite{chen2015microsoft}) and a  video-text retrieval dataset~(MSRVTT~\cite{xu2016msr}), referred as \textbf{\textit{downstream task datasets (DTD)}} in this paper. 
 Target-language captions are obtained by automatically translating the English captions in DTD with Google Translate. 
 Besides, the web-scraped image-caption dataset CC3M~\cite{sharma2018conceptual} with machine-translated captions is also used for training, from which 300k image-captions pairs are randomly selected and known as CC300K~\cite{zhang2022multi}.

\subsubsection{Retrieval Settings.} 
% For CCR, we evaluate the performance of our DASD against SOTA approaches under two settings:
We conduct experiments under two CCR settings:
(1) \textit{\textbf{Cross-lingual Finetune}}: we first train models using English data in DTD and then further finetune models with target-language data produced by MT tools. Finally, models are tested on DTD target-language datasets.
(2) \textit{\textbf{Zero-shot}}: models are trained on commonly-used datasets~(e.g., CC300K) and then directly evaluated on DTD without any DTD finetuning.
%The results under both settings are summarized in Table~\ref{table1}.

\subsubsection{Evaluation Metrics.} 
For image-text retrieval, following~\cite{zhang2022multi}, we report the mean Average Recall (mAR) for image-text retrieval.
For video-text retrieval, we follow~\cite{rouditchenko2023c2kd} and use text→video Recall@1 score to evaluate the performance.

\input{table_image_text_sota}
\subsection{Evaluation on Cross-lingual Image-Text Retrieval}

%represented it in the table~\ref{table1} using the DTD format.
%we fine-tune the VLP model on downstream English data, then freeze the parameters of the VLP model and train the adapter on non-English data. 
%Please note that this does not violate the CCR task setting, as machine-translated data is allowed to be used.

Under the \textit{\textbf{Cross-lingual Finetune}} setting, image-caption pairs for target languages are obtained in two separate ways: (1) we directly leverage the target-language data in CC300K (following MLA~\cite{zhang2022multi}). 
(2) English captions in Multi30K and MSCOCO are converted into target languages utilizing Google Translate (following CL2CM~\cite{wang2024cl2cm}).
In both cases, it can be seen in Table~\ref{table1} that DASD outperforms all the comparison methods, demonstrating the effectiveness of %leveraging disentangled semantics to generate
dynamic adapters.
Please note that the SOTA model CL2CM relies on the full-model training, besides, its cross-lingual alignments is carefully designed where token-level alignments are involved to improve the final performance, all of which consumes a large amount of computing budgets.
Although DCOT and CL2CM are not open-sourced, they all expand the cross-attention module in NRCCR and therefore are expected  to have more trainable parameters than NRCCR.
%For a fair comparison, we used the same datasets for training as in previous work. We employed two datasets: CC300K and non-English text-image pairs from the downstream task dataset, demonstrating the effectiveness of leveraging disentangled semantics to generate dynamic adapters..
%Additionally, to explore the potential of DASA, we conducted an experiment using both datasets together. 
%Our experimental process followed the work of~\cite{zhang2022multi}, dividing the training into two stages: the first phase for cross-lingual alignment and the second phase for cross-modal alignment.
%By comparison, it can be seen that our method outperforms all the comparison methods, demonstrating the effectiveness of leveraging disentangled semantics to generate dynamic adapters.
Under the \textbf{\textit{Zero-shot}} setting, 
we observe that the performances of NRCCR, DCOT and CL2CM drop severely due the absence of downstream datasets, which are surpassed by the parameter-efficient model MLA. 
Among them, DCOT~\cite{wang2024dual} tries to learn noisy correspondence in CCR by quantifying the confidence of the sample pair correlation with optimal transport theory from both the cross-lingual and cross-modal views. Compared with  PEFT models, DCOT relies on full-model training, rendering their method much more time and computing consuming. Besides, DCOT only focus on reducing the impact of obvious errors brought by MT, neglecting other factors (e.g., language gaps) which could also hurt the cross-lingual alignment and result in degraded performance.
Our performance still achieves the best performance when downstream task data is not available, showing strong zero-shot cross-lingual transfer ability.
%Although MLA also uses CC300K as training data, which appears to be fair, we believe that images with distribution bias limit our potential. Therefore, we included the MT downstream task dataset in our finetune-en experiment to demonstrate our performance.

\input{table_video_text_sota}
\subsection{Evaluation on Cross-lingual Video-Text Retrieval}
For cross-lingual video-text retrieval, experiments are conducted on MSRVTT~\cite{xu2016msr} under the same settings with cross-lingual image-text retrieval, where the model searches for the most semantically relevant videos given a text query in a low-resource language.
%For video encoding, following the prior work~\cite{luo2021clip4clip}, we first uniformly sample 12 frames from the video and use a pre-trained visual encoder to extract representations for each frame.
%Then, the video representation is obtained by performing average pooling over the frame representations.
We report the text$\rightarrow$video Recall@1 score in Table~\ref{vtt}.
%We expanded the number of languages in CC300K to match those in the Multi-MSRVTT dataset using Google Translate. 
%Under the \textbf{\textit{Zero-shot}} setting, models are finetuned on machine-translated CC300K without access to any video data. 
%Under the \textbf{\textit{Cross-lingual Fine-tune}} setting, text-video pairs from MSRVTT and their target-language counterparts generated by MT are employed to further improve the model performances.
Under both the \textbf{\textit{Zero-shot}} and \textbf{\textit{Cross-lingual Finetune}} settings, 
we simply adopt the same hyperparameter values and training strategy used for the cross-lingual image-text retrieval. 
%the performance of DASD is slightly lower than MLA on the source language (English).
As shown in Table~\ref{vtt}, for both settings, our model consistently outperforms MLA on all eight target languages, demonstrating a strong cross-lingual ability for text-video retrieval.

\input{table_compatilitity}
\subsection{Generalizability Analysis}
Since the adapter is a lightweight, plug-and-play module, we also  investigate whether our proposed DASD is compatible with different VLP models. 
To this end, we substitute the pretrained CLIP with the recently-proposed M-VLP model CCLM~\cite{zeng2023cclm}, which has been  pretrained on the combination of image-caption pairs and parallel corpora.
Specifically, since CCLM is a single-stream model and difficult to extend directly, we follow~\cite{wang2024cl2cm} to modify CCLM into a dual-stream model and apply DASD to its text encoder.
As reported in Table~\ref{transfer},
the cross-lingual text-image retrieval performance of CCLM is further improved when equipped with our proposed dynamic adapters, outperforming the framework built upon CLIP.
%demonstrating the good compatibility of our method.
These results verify that our method is compatible with various VLP models and could achieve a higher performance when equipped with stronger VLP models.

\subsection{Ablation Studies}
To verify the effectiveness of each component in DASD,
we conduct ablation studies under the cross-lingual finetune setting on  Multi30K and MSCOCO.
%Considering the setting of CCR, target-language sentences used for training are all obtained by machine-translating the English captions in Multi30K and MSCOCO. 

\input{table_ablation_wz}
\subsubsection{The effectiveness of input-conditional parameters.}
We first investigate the contribution of the dynamic parameters 
by removing the input-conditional parameter matrix $W^z$ out of DASD, turning it into the traditional adapter with only static parameters.
As reported in Table~\ref{tab:wz}, when using static adapters for cross-lingual transfer, we observe a severe performance degradation on all five target languages,
demonstrating the importance of the dynamic parameters.

\input{table_ablation_modules}
\subsubsection{The effectiveness of semantic-related and semantic-agnostic features.}
We then study the effectiveness of $f^{sa}$ and $f^{sr}$ to dynamic parameter generation.
As summarized in Table~\ref{tab:modules},  the inclusion of both kinds of features leads to a certain improvement in mAR on all five target languages.
In the case where only one kind of features is employed, we observe that the semantic-related features having a slightly greater positive impact compared to the semantic-agnostic ones. 
It not only demonstrates the effectiveness of both kinds of features, but also shows the
complementary of the semantic-related and semantic-agnostic features.
%When both  kinds of features are removed, the parameter matrix $W^z$ is unavailable, turning our dynamic adapters into the traditional adapters.
%In this case, as shown in the last row of  Table~\ref{tab:modules}, the sum mAR of the model drops significantly
%from 446.3 to 429.4, verifying the importance of dynamic parameters.

\input{table_ablation_losses}
\subsubsection{The impact of semantics disentangling.}
To examine the necessity of performing semantics disentangling, we investigate the impact brought by different disentangling losses~($\mathcal{L}_{adv}$ and $\mathcal{L}_{sc}$) and report results in Table~\ref{tab:loss}. 
We observe that with the loss constraints added, the model performance increases on all five languages, validating the effectiveness of the adversarial training and semantic distillation.
To our knowledge, no prior work has applied dynamic adapters or semantics disentangling to CCR, and our work fills this gap and thereby gains a clear improvement.

\subsection{Visualization Analysis}
\begin{figure}[tb!]
\centering
\includegraphics[width=0.95\columnwidth]{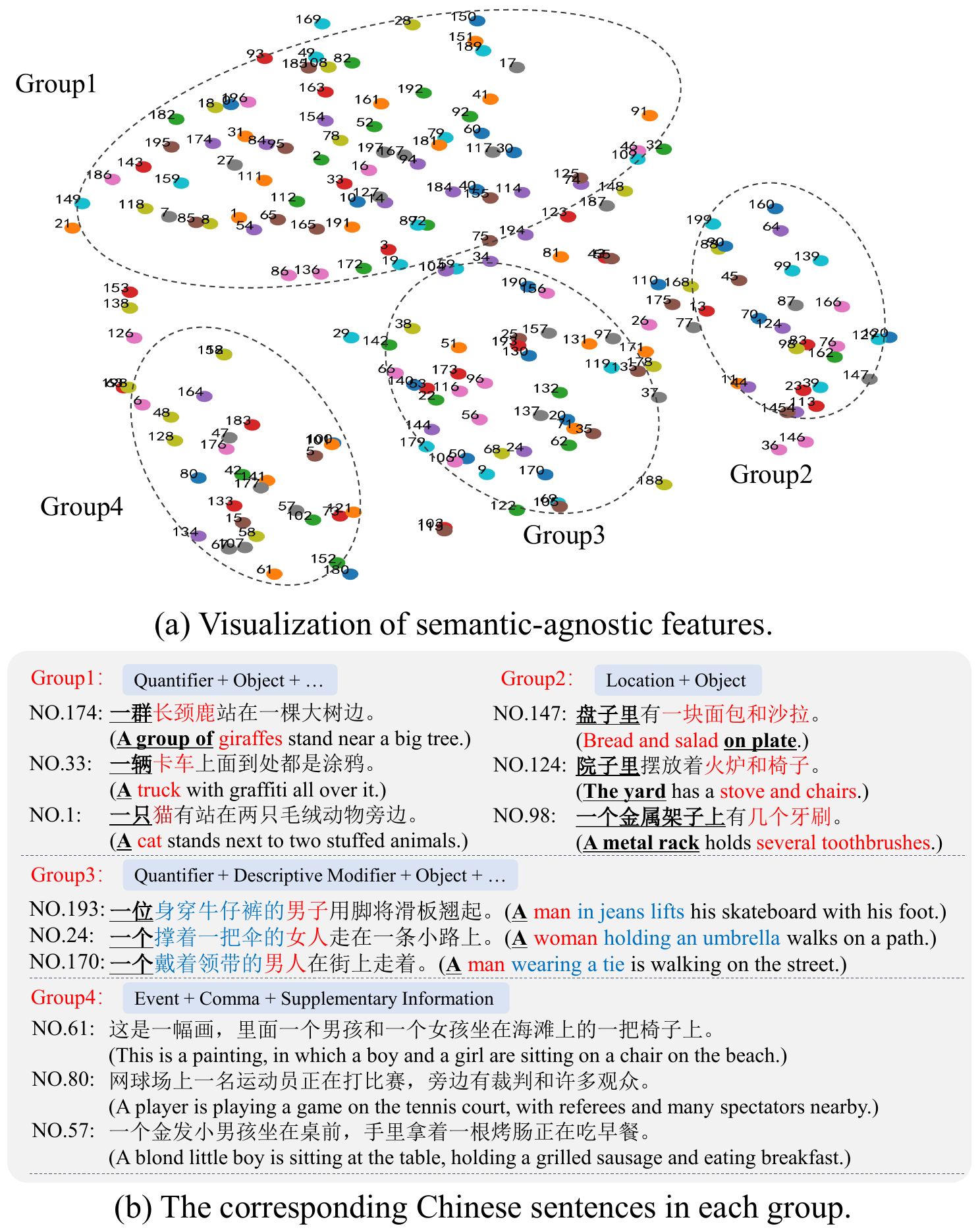}
\vspace{-3mm}
\caption{Visualization of the semantic-agnostic features extracted from 200 randomly-selected Chinese sentences in MSCOCO testset.} 
\label{visual}
\end{figure}

In Figure~\ref{visual}, we use t-SNE to visualize the semantic-agnostic representations of 200 Chinese sentences randomly selected from MSCOCO testset. 
As illustrated in Figure~\ref{visual}(a), the semantic-agnostic representations produced by DASD have been automatically clustered into 4 groups. 
Figure~\ref{visual}(b) lists some corresponding sentences of each group, and we observe that sentences in the same group are expressed in a similar way.
Concretely, sentences in group 1 start with a quantifier followed by the key object,
and sentences in group 2 begin with the location of the key object. 
%Group 3 describes a certain kind of person is doing something. 
Compared to group 1, sentences in group 3 include an additional descriptive modifier before the key object. 
Group 4 contains long sentences that are divided into two parts. 
This visualization result confirms that our DASD effectively captures semantic-agnostic characteristics of captions through semantics disentangling.

%Group 4 describes what kind of person is doing something. Compared to Group 1, it includes an additional descriptive modifier, and the subject has changed to a person.
%Group 1 describes what an object is doing or what an object looks like.
%Group 2 sentences all begin with positional words, introducing the location or place described in the sentence.
%Group 3 contains long sentences that can be divided into multiple parts.
%Group 4 describes what kind of person is doing something. Compared to Group 1, it includes an additional descriptive modifier, and the subject has changed to a person.
%Thus, we can see that the semantic-agnostic representations focuses on the structural characteristics of the sentences.

% \captionsetup[subfloat]{font=small} 
% \begin{figure}[tb!]
% \centering
% % \vspace{-2mm}
% \subfloat[Visualization of semantic-agnostic features.]
% {\includegraphics[width=0.40\textwidth]{images/tsne_part1.pdf}}%
% \hfil
% \vspace{-4mm}
% \subfloat[The corresponding Chinese sentences in each group.]
% {\includegraphics[width=0.45\textwidth]{images/tsne_part2.pdf}}%
% \vspace{-3mm}
% \caption{Visualization of 200 random Chinese sentences from machine-translated MSCOCO.}
% \vspace{-6mm}
% \label{visual}
% \end{figure}

%% file: table_image_text_sota.tex
\begin{table*}[ht!]
\caption{Cross-lingual image-text retrieval results on Multi30K and MSCOCO. \#TP: \textit{the number of Trainable parameters}, DTD: \textit{Down-stream Task Datasets} (i.e., Multi30K and MSCOCO).
Despite being equipped the dynamic parameter generator, the number of trainable parameters  in DASD remains comparable with MLA using the static adaptor, maintaining the parameter efficiency of the model while achieving significant improvements under both settings.}
\vspace{-3mm}
\centering

\renewcommand{\arraystretch}{1.2}
% \scalebox{0.9}{
\small
\begin{tabular}{llcccllccccc}
\hline
 & \multirow{2}{*}{\textbf{Method}}   & \multirow{2}{*}{\textbf{\#TP}} & \multicolumn{2}{c}{\textbf{Training Data}} & & \multicolumn{3}{c}{\textbf{Multi30K}}                                                                        &  & \multicolumn{2}{l}{\textbf{MSCOCO}} \\ \cline{4-5} \cline{7-9} \cline{11-12}
 &                 &                 & \textbf{English}        & \textbf{Target Languages}                     &         & \multicolumn{1}{c}{\textbf{DE}} & \multicolumn{1}{c}{\textbf{FR}} & \multicolumn{1}{c}{\textbf{CS}} & & \multicolumn{1}{c}{\textbf{ZH}}   & \multicolumn{1}{c}{\textbf{JA}}   \\ \hline
\multirow{8}{*}{\rotatebox{90}{\textit{\textbf{Cross-lingual Finetune}}}} 
     % & M3P~\cite{ni2021m3p}         & 566M    & CC3M+Wiki           & 82.1               & 67.3                                & 65.0                &  & 55.5             & 56.0             \\
     & UC$^2$~\cite{zhou2021uc2}       & 478M    & DTD & CC3M     &            & 83.8               & 77.6                                & 74.2                &  & 82.0             & 71.7             \\
     & MURAL~\cite{jain2021mural}   & 300M    & DTD & CC300K   &            & 76.5               & 76.7                                & 70.1                &  & -                & 74.6             \\
     & MLA~\cite{zhang2022multi}    & 108M    & DTD & CC300K   &      &  86.4      & 87.3                        & 79.5        &  & -                & 80.4      \\
     & DASD (ours)                  & 134M    & DTD & CC300K   &           & \textbf{87.4}      & \textbf{88.6}                       & \textbf{83.4}       &  & \textbf{88.5}    & \textbf{84.8}     \\ \cline{2-12} 
     & NRCCR~\cite{wang2022cross}   & 216M    & DTD & MT(DTD)     &             & 80.1               & 80.4                                & 77.9                &  & 85.4             & 84.5             \\
     & DCOT~\cite{wang2024dual}     & -       & DTD &MT(DTD)     &             & 82.5               & 82.6                                & 80.3                &  & 86.9             & 85.9             \\
     & CL2CM~\cite{wang2024cl2cm}   & -       & DTD & MT(DTD)     &             & 83.0       & 83.3                       & 80.9        &  & 87.0    & 86.0      \\    
     & DASD (ours)                  & 134M    & DTD & MT(DTD)     &             & \textbf{88.5}      & \textbf{91.1}                       & \textbf{87.6}                &  & \textbf{90.0}      & \textbf{89.1}      \\\hline
\multirow{8}{*}{\rotatebox{90}{\textit{\textbf{Zero-shot}}}}   
     % & M3P~\cite{ni2021m3p}         & 566M    & CC3M+Wiki            & 36.8                & 27.1                               & 20.4                &  & -                & 33.3             \\
     & UC$^2$~\cite{zhou2021uc2}       & 478M    & -       & CC3M            &     & 62.5                & 60.4                               & 55.1                &  & -                & 62.3             \\
     & MURAL~\cite{jain2021mural}   & 300M    & - & CC300K     &          & 62.7                & 60.8                               & 57.5                &  & -                & 62.5            \\
    & MLA~\cite{zhang2022multi}     & 108M    & - & CC300K     &          & 80.8        & 80.9                       & 72.9       &  & 78.5     & 76.7       \\
     & DASD (ours)                  & 134M    & - & CC300K     &        &\textbf{81.9}        & \textbf{82.1}                      &\textbf{74.3}        &  & \textbf{79.6}    & \textbf{77.5}       \\ \cline{2-12} 
     & NRCCR~\cite{wang2022cross}   & 216M    & - & MT(MSCOCO)    &          & 74.8                & 72.3                               & 68.5                &  & -                &  -                \\
     & DCOT~\cite{wang2024dual}     & -       & - & MT(MSCOCO)   &           & 76.5                & 74.2                               & 70.7                &  & -             & -                   \\
     & CL2CM~\cite{wang2024cl2cm}   & -       & - & MT(MSCOCO)    &          & 76.9         & 74.5                       &  71.5       &  & -                & -                 \\
     & DASD (ours)                  & 134M   & - & MT(MSCOCO)    &          &\textbf{80.1}        & \textbf{81.3}                      &\textbf{74.9}        &  & -    & -       \\ \hline
\end{tabular}

% \vspace{-3mm}

\label{table1}
\end{table*}

%% file: table_video_text_sota.tex
\begin{table*}[tb!]
\caption{Cross-lingual video-text retrieval results on Multi-MSRVTT, CL-FT: \textit{Cross-lingual Fine-tune}, ZS: \textit{Zero-shot}.
Our proposed DASD performs the best over baseline methods on all target languages.}
% Best results are in \textbf{bolded} and second best are \underline{underlined}.}
\renewcommand{\arraystretch}{1.2}
\vspace{-3mm}
\centering
% \scalebox{0.85}{
\small
\begin{tabular}{llccccccccc}
\hline
% & \textbf{Method} & en & de & fr & cs   & zh   & ru   & vi   & sw   & es   & avg  \\ \hline
% \multirow{3}{*}{\rotatebox{90}{ZS}}
%     & MMP~\cite{huang2021multilingual}             & 23.8             & 19.4            & 20.7          & 19.3         & 18.2              & 19.1          & 8.2          & 8.4               & 20.4               & 17.5 \\
%     & MLA~\cite{zhang2022multi}                    & \textbf{30.8}    & 20.1    & 22.0  & 15.7                 & 18.3      & 14.4                  & 8.2          & 10.7      & 20.2                       & 17.8 \\
%     & DASD (ours)                                          & 28.7     & \textbf{23.7}   & \textbf{23.9} & \textbf{21.4}        & \textbf{22.4}     & \textbf{21.7}         & \textbf{11.2}        & \textbf{15.3}     & \textbf{23.1}              & \textbf{21.3} \\ \hline
& \textbf{Method}                   & \textbf{DE}          & \textbf{FR}          & \textbf{CS}   & \textbf{ZH}   & \textbf{RU}   & \textbf{VI}   & \textbf{SW}   & \textbf{ES}   & \textbf{SUM}  \\ \hline
\multirow{4}{*}{\rotatebox{90}{CL-FT}}     
    & MMP~\cite{huang2021multilingual}                          & 21.1            & 21.8           & 20.7                 & 20.0             & 20.5         
            & 10.9                 & 14.4              & 21.9                       & 151.3%19.4 
            \\
    & C2KD~\cite{rouditchenko2023c2kd}             & 24.7            & 25.4           & 24.0         & 23.4             & 23.1 
            & 13.6        & 20.3      & 25.5                       & 180.0%22.9
            \\
    & MLA~\cite{zhang2022multi}                   & 26.1    & 26.7   & 20.5                 & 25.3     & 18.9         
            & 12.9                 & 12.6              & 27.2              & 170.2%23.6
            \\
    & DASD (ours)                                          & \textbf{28.8}   & \textbf{30.5}  & \textbf{26.3}        & \textbf{28.0}    & \textbf{25.9} 
            & \textbf{14.8}        & \textbf{22.1}     & \textbf{29.7}              & \textbf{206.1}%\textbf{27.3} 
            \\ \hline
%& \textbf{Method} & \textbf{EN} & \textbf{DE} & \textbf{FR} & \textbf{CS}   & \textbf{ZH}   & \textbf{RU}   & \textbf{VI}   & \textbf{SW}   & \textbf{ES}   & \textbf{AVG}  \\ \hline
\multirow{3}{*}{\rotatebox{90}{ZS}}
    & MMP~\cite{huang2021multilingual}                         & 19.4            & 20.7          & 19.3         & 18.2              & 19.1          & 8.2          & 8.4               & 20.4               & 133.7%17.5 
    \\
    & MLA~\cite{zhang2022multi}                       & 20.1    & 22.0  & 15.7                 & 18.3      & 14.4                  & 8.2          & 10.7      & 20.2                       & 129.6%17.8
    \\
    & DASD (ours)                                          & \textbf{23.7}   & \textbf{23.9} & \textbf{21.4}        & \textbf{22.4}     & \textbf{21.7}         & \textbf{11.2}        & \textbf{15.3}     & \textbf{23.1}              & \textbf{162.7}%\textbf{21.3}
    \\ \hline
\end{tabular}

\vspace{-2mm}
\label{vtt}
\end{table*}

%% file: table_compatilitity.tex
\begin{table}[tb!]
\caption{The performances of our method using different VLP models as the backbone.
Our dynamic adapter could not only expand the monolingual VLP model (CLIP) to multiple target languages, but also exhibits a good compatibility with different VLP models.}
\vspace{-3mm}
\centering
\setlength{\tabcolsep}{0.7mm}{
\renewcommand{\arraystretch}{1.2}
% \scalebox{0.83}{
\small
\begin{tabular}{lccclccc}
\hline
\multicolumn{1}{l}{\multirow{2}{*}{\textbf{Method}}} & \multicolumn{3}{c}{\textbf{Multi30K}}                  &  & \multicolumn{2}{c}{\textbf{MSCOCO}} 
&\multicolumn{1}{c}{\multirow{2}{*}{\textbf{SUM}}} \\ \cline{2-4} \cline{6-7} 
\multicolumn{1}{c}{}  & \textbf{FR}  & \textbf{DE} & \textbf{CS} &  & \textbf{ZH} & \textbf{JA}  \\ \hline
CLIP~\cite{radford2021learning} & - & - & - &  & - & - & -   \\
CLIP+ours  & 91.1 & 88.5& 87.6&  & 90.0& 89.1 & 446.3   \\
\hline
CCLM~\cite{zeng2023cclm}   & 81.7& 83.9 & 80.2  &  & 85.2 & 82.7  & 413.7       \\
%CCLM+DCOT & 86.4 & 85.8  & 85.3  &  & 89.2 & 89.3 \\
%CCLM+CL2CM & 89.3 & 88.4  & 87.7 &  & 90.7   & 91.0          \\
CCLM+ours  & \textbf{91.4} & \textbf{89.3} & \textbf{88.5} &  & \textbf{91.9} & \textbf{91.6} & \textbf{452.7} \\ \hline

\end{tabular}}

% \vspace{-3mm}

\label{transfer}
\end{table}

%% file: table_ablation_wz.tex
\begin{table}[tb!]
\caption{Effectiveness of the dynamic adapter for CCR on Multi30K and MSCOCO.
Using input-conditional parameters~($W^z$)  brings in substantial performance gain.}
\vspace{-2mm}
% \centering
\setlength{\tabcolsep}{0.9mm}{
\renewcommand{\arraystretch}{1.2}
% \scalebox{0.83}{
\small
\begin{tabular}{lccclccc}
\hline
\multicolumn{1}{l}{\multirow{2}{*}{\textbf{Method}}} & \multicolumn{3}{c}{\textbf{Multi30K}}                  &  & \multicolumn{2}{c}{\textbf{MSCOCO}} 
&\multicolumn{1}{c}{\multirow{2}{*}{\textbf{SUM}}} \\ \cline{2-4} \cline{6-7} 
\multicolumn{1}{c}{}  & \textbf{FR}  & \textbf{DE} & \textbf{CS} &  & \textbf{ZH} & \textbf{JA}  \\ \hline
Traditional Adapter   & 88.3 & 87.0 & 83.1 &  & 85.3 & 85.7 & 429.4 \\
%CCLM+DCOT & 86.4 & 85.8  & 85.3  &  & 89.2 & 89.3 \\
%CCLM+CL2CM & 89.3 & 88.4  & 87.7 &  & 90.7   & 91.0          \\
Dynamic Adapter (ours)  & \textbf{91.1} & \textbf{88.5}& \textbf{87.6}&  & \textbf{90.0} & \textbf{89.1}  & \textbf{446.3}   \\ \hline
\end{tabular}}

% \vspace{-3mm}
\label{tab:wz}
\end{table}

%% file: table_ablation_modules.tex
\begin{table}[tb!]
\caption{Effectiveness of the semantic-related and semantic-agnostics features ($f^{sr}$ and $f^{sa}$).}
%In the last row dynamic parameter\caption{Effectiveness of the semantic-related and semantic-agnostics features ($r^{sr}$ and $r^{sa}$).}s are not available due to the absence of both kinds of features, turning our DASD into traditional adapters.}
\vspace{-3mm}
\centering
\setlength{\tabcolsep}{1.5mm}{
\renewcommand{\arraystretch}{1.2}

% \scalebox{0.8}{
\small
\begin{tabular}{cccccccccc}
\hline
\multicolumn{2}{c}{\textbf{Features}}  &  & \multicolumn{3}{c}{\textbf{Multi30K}} &  & \multicolumn{2}{c}{\textbf{MSCOCO}} & \multirow{2}{*}{\textbf{SUM}} \\
\cline{1-2} \cline{4-6} \cline{8-9}
$f^{sr}$  & $f^{sa}$   && \textbf{FR}& \textbf{DE} & \textbf{CS} &  & \textbf{ZH} & \textbf{JA}   &  \\ \hline
$\checkmark$ & $\checkmark$ &  & \textbf{91.1} & \textbf{88.5}& \textbf{87.6}&  & \textbf{90.0} & \textbf{89.1}  & \textbf{446.3}   \\
$\checkmark$& $\times$&  & 90.2  & 87.7 & 86.8  &  & 89.1 & 88.0  & 441.8                           \\
$\times$ & $\checkmark$&  & 89.9 & 87.8 & 86.3 &  & 88.7 & 88.1& 440.8 \\ 
%$\times$  & $\times$ & & 88.3 & 87.0 & 83.1 &  & 85.3 & 85.7 & 429.4   \\

\hline
\end{tabular}}
% \vspace{-3mm}
\label{tab:modules}
\end{table}

%% file: table_ablation_losses.tex
\begin{table}[tb!]
\caption{Effectiveness of different disentangling losses.}
\vspace{-3mm}
\centering
\setlength{\tabcolsep}{1.5mm}{
\renewcommand{\arraystretch}{1.2}
\small
% \scalebox{0.8}{
\begin{tabular}{cccccccccc}
\hline
\multicolumn{2}{c}{\textbf{loss}} &  & \multicolumn{3}{c}{\textbf{Multi30K}}         &           & \multicolumn{2}{c}{\textbf{MSCOCO}} & \multirow{2}{*}{\textbf{SUM}} \\ \cline{1-2} \cline{4-6} \cline{8-9}
\textbf{$\mathcal{L}_{sc}$}   & \textbf{$\mathcal{L}_{adv}$}  &  & \textbf{FR}  & \textbf{DE}  & \textbf{CS} &   & \textbf{ZH} & \textbf{JA} &                               \\ \hline
$\checkmark$  & $\checkmark$  &  & \textbf{91.1} & \textbf{88.5} & \textbf{87.6} & \textbf{} & \textbf{90.0}    & \textbf{89.1}    & \textbf{446.3}                             \\
$\checkmark$   & $\times$   &  & 90.7  & 87.9  & 87.3  & & 89.6 & 88.3 & 443.8        \\
$\times$   & $\checkmark$ &  & 90.6 & 88.1& 86.9& & 89.5& 88.5& 443.6     \\
$\times$ & $\times$ &  & 90.1 & 87.4 & 86.5 &  & 89.1  & 88.0  & 441.1  \\ \hline
\end{tabular}
}

\label{tab:loss}
\end{table}

%% file: conlusion.tex
\section{Conclusion}
This paper proposes dynamic adapters with semantics disentangling for CCR. 
By characterizing target-language captions from two distinct and complementary aspects, our DASD dynamically generates adapters for input captions in varied forms. 
Extensive experiments show the effectiveness of DASD and its new SOTA performance. 
Given DASD is simple and effective, we believe it can also be used as a new strong baseline for other cross-lingual transfer tasks.

%% file: acknowledgement.tex
\section{Acknowledgments}
This work was supported by the Pioneer and Leading Goose R\&D Program of Zhejiang (No. 2024C01110), National Natural Science Foundation of China (No. 62306278), Zhejiang Provincial Natural Science Foundation (No. LZ23F020004 and No. LQ23F020008), Young Elite Scientists Sponsorship Program by China Association for Science and Technology (No. 2022QNRC001),  Fundamental Research Funds for the Provincial Universities of Zhejiang~(No. FR2402ZD).

%% file: supplementary.tex
\section{Supplementary Material}

This supplementary material contains the following contents which are not included in the paper due to space limits: 
\begin{itemize}
    \item Detailed descriptions of the datasets used in our experiments.
    \item Implementation details including computation cost, model structure and training details.
    \item Additional ablation studies about the influence of SDM backbones and  dynamic adapter size.
    \item The t-SNE visualization of learned semantic-related and semantic-agnostic representations.
\end{itemize}

\subsection{Datasets}\label{sec:dataset}
Evaluations are performed on two image-text retrieval datasets (Multi30K~\cite{elliott2016multi30k} and MSCOCO~\cite{chen2015microsoft}) and a  video-text retrieval dataset~(MSRVTT~\cite{xu2016msr}), referred to as \textit{Downstream Task Datasts}~(DTD) in the main text. 
 Target-language captions are obtained by automatically translating the English captions in DTD with Google Translate. 
 Besides, the web-scraped image-caption dataset CC3M~\cite{sharma2018conceptual} expanded with machine-translated captions~\cite{zhou2021uc2} is also used for training, from which 300k image-captions pairs are randomly selected and known as CC300K~\cite{zhang2022multi}.
\begin{itemize}

\item \textbf{Multi30K}~\cite{elliott2016multi30k}: This dataset is built by extending Flickr30K~\cite{young2014image} from English to German, French and Czech. It consists of 31K images, each paired with 5 captions in English and German, and 1 caption in French and Czech. 
We split the dataset following~\cite{young2014image}. 

\item \textbf{MSCOCO}~\cite{chen2015microsoft}:  The original dataset consists of 123,287 images, and each image is annotated with 5 English captions. 
Previous works further add 5 Japanese captions for all images~\cite{yoshikawa2017stair} and 1 Chinese captions for 20,000 images~\cite{li2019coco}.
We split the dataset following~\cite{zhou2021uc2}.

\item \textbf{MSRVTT}~\cite{xu2016msr}: The original dataset is a video-caption dataset containing 10,000 videos, each with 20 English captions. 
We use its multilingual version~\cite{huang2021multilingual}, in which the English captions are translated to 8 languages~(German, French, Russian, Spanish, Czech, Swahili, Chinese and Vietnamese) with MT.
We split the dataset following~\cite{huang2021multilingual}.
%We follow the same setup from the previous work~\cite{huang2021multilingual} for our experiments.

\item \textbf{CC3M\&CC300K}: CC3M~\cite{sharma2018conceptual} dataset contains 3.3 million English image-text pairs scraped from the web, which is further expanded by UC$^2$~\cite{zhou2021uc2} to  5 languages (German, French, Czech, Chinese and Japanese) with the aid of MT tools;
The recent work MLA~\cite{zhang2022multi} randomly selects 300K image-text pairs from CC3M and then converts them into target languages using MT tools, referred to as \textbf{CC300K}.
%to train our model when downstream datasets are not available, demonstrating the low-cost advantage of DASD. 

\end{itemize}


% \captionsetup[subfloat]{font=small} 
% \begin{figure}[tb!]
% \centering
% % \vspace{-2mm}
% \subfloat[Visualization of semantic-agnostic features.]
% {\includegraphics[width=0.40\textwidth]{fig/tsne_part1.pdf}}%
% \hfil
% \vspace{-2mm}
% \subfloat[Text Content Display]
% {\includegraphics[width=0.45\textwidth]{fig/tsne_part2.pdf}}%
% \vspace{-2mm}
% \caption{Visualization of 200 Sentences on the MSCOCO.}
% \vspace{-3mm}
% \label{visual}
% \end{figure}

\subsection{Implementation Details}\label{sec:details}
In this section, we describe the implementation details. 
Our source code is anonymously  released  at \url{https://github.com/HuiGuanLab/DASD} so that readers of interest can have full access to every implementation detail.

\textit{Computation\&Time Cost.}
We are conduct experiments using a single RTX 3090 GPU with the PyTorch 1.10.1 framework on a Linux system equipped with 128GB of memory.
As for the cross-lingual transfer to all five target languages in Multi30K and MSCOCO, it takes no more than 4 hours and 9 hours to train the model under the zero-shot and cross-lingual finetune settings, respectively.

\textit{Model Structure.}
The backbone of our DASD is CLIP-ViT-B~\cite{radford2021learning} whose parameters stay frozen during the cross-lingual transfer. 
The image encoder of the pretrained CLIP~\cite{radford2021learning} has been aligned with the English text encoder through contrastive learning on 400M English image-text pairs.
The base version of multilingual BERT~(mBERT, \cite{devlin2018bert}) is utilized as the embedding block for all target languages.
The trainable MLP in semantic-related and semantic-agnostic modules only have one hidden layer whose size is set to 256.
In the target-language branch $\Phi^T$, the hidden-layer size $d_u$ in the dynamic adapters is set to 32.

\textit{Training Details.}
The training batch size is 128.
The value of temperature coefficient in Equation 15 (in
the main text) is set to 0.01.
For loss coefficients in Equation 16 (in
the main text), we set $\lambda_1 = 1$ and $\lambda_2 = 0.1$. 
For cross-lingual alignment, 45,000 steps are performed with a learning rate of 2e-4.
For cross-modal alignment,  6,000 steps are performed with a learning rate of 6e-6.  
For adversarial training, the discriminator $F$ is trained simultaneously with other parts of DASD, whose parameters are updated with Adam optimizer using the learning rate 2e-4 during the cross-lingual alignment stage and the learning rate 6e-6 during the cross-modal alignment stage.
During training, we use the Adam optimizer with a linear warm-up for the first ${10\%}$ of steps.
Under the cross-lingual finetune setting, we first optimize the English text encoder and the image encoder on the down-stream task dataset for 5 epochs with a learning rate of 3e-6, and then proceed with cross-lingual and cross-modal alignment.

\subsection{Additional Ablation Studies}\label{sec:klayers}

\subsubsection{The Influence of Semantic Consistency Loss}
We first investigate the influence caused by the use of different loss in semantic consistency.
In the main text Equation 14, $L1$ distance is used to define the semantic consistency loss. 
Here we try other losses such as Smooth $L1$ and $L2$, and experimental results on Multi30k and MSCOCO are shown in Table~\ref{con_loss}, which verifies that the loss $L1$ works better than others.
\input{table_ablation_LossFunction}

\subsubsection{The Influence of Training Data Size}
Both CC300K and CC3M are used for the experiments in the main text, while the latter is ten times the size of the former. In Table 1 (in the main text), the proposed model only uses CC300K for training, here we investigate how much performance gain could be gained by using the larger training dataset (CC3M).
Specifically, we train our model using CC300k and CC3M separately and then perform evaluations on Muli30K and MSCOCO. 
As shown in Table~\ref{table_3M}, the larger dataset brings slight improvements in all five target languages, demonstrating that our model can be further improved when more training data are used. 
\input{table_ablation_CC3M}

\subsubsection{The Influence of CLIP Layers}
For semantic disentanglement, only the first layer of CLIP is utilized to encode captions, with no performance degradation compared to using 12 layers. 
In order to examine whether all 12 layers in the target-language branch are necessary,  we conduct experiments with less CLIP layers on Multi30k and MSCOCO. 
As shown in Table~\ref{clip_layers}, when using only the bottom layer of CLIP, the performance of our model drops severely on all five target languages. 
With more CLIP layers employed, the model performance increases steady and reaches its best when using 12 layers, verifying the necessity of our current model scale.
\input{table_ablation_drop_ClipLayer}

\subsubsection{The Influence of Dynamic Adapters}
In our DASD, the dynamic adapters are distributed at 12 CLIP layers, so it is necessary to investigate whether these dynamic adapters contribute equally at each layer.
Specifically, we investigate the contribution of dynamic adapters  in each layer by separately removing them out of DASD. 
As shown in Table~\ref{table_DA}, the performance of our model degrades when DA of a certain layer are removed, verifying the necessity of all DA in our model.
We also observe that removing DA of the top layer leads to a more severe performance drops than other layers, suggesting that top-layer DA play a more important role than others.
\input{table_ablation_woDA}
\input{table_ablation_layers}
\subsubsection{The Influence of SDM Backbones}
Recall that the backbone of our Semantic Disentangling Module (SDM) is only the first pretrained layer of CLIP whose parameters are frozen.
%In this experiment, we examine whether DASD can achieve its optimal performance with only one pre-trained layer used for semantic disentangling.
Figure~\ref{fig:Klayer} shows how the mAR of our model varies with  the number of transformer layers employed in SDM. 
%For example, 1+1 indicates that the semantic-related branch uses only one pre-trained model layer and the semantic-unrelated branch also uses only one layer.
When layer number is 0, no pretrained transformer layer is used in SDM, instead, we employ two trainable MLP modules to extract semantic-related and semantic agnostic features, respectively.
%which shares the same size with the adapter $\mathtt{A^{sr}}$ and $\mathtt{A^{sa}}$. 
%This way, the number of trainable parameters for 0 layers and 1 layer is exactly the same. 
%Through comparison, it becomes clear that having one layer of pretrained transformer is quite necessary. 
Interestingly, we observe a large jump in performance when only one pretrained layer is employed.
With more pretrained layers stacked,  the mAR score stays above 89.0 with slight fluctuations, indicating that using only one pretrained transformer as the backbone is an optimal choice for SDM.

\subsubsection{The Influence of Dynamic Adapter Capacity}
%In the main text, we describe how we used the reshape operation to convert semantic-related and semantic-agnostic features into matrix form.
The keep the parameter efficiency of DASD, the hidden-layer dimension ($d_u$)
in dynamic adapters is much smaller than the prior work~\cite{zhang2022multi}.
In this experiment, we investigate the influence brought by the capacity of dynamic adapters. 
%We also conducted ablation experiments to evaluate the impact of different values for $d_u$. 
Table~\ref{tab:hidden_size} reports the performance of our model with various hidden-layer sizes $d_u$.
When $d_u$ increases from 16 to 32, we observe a clear performance gain on all five target languages. 
However, the improvement becomes insignificant when $d_u$ is further increased to 64, verifying that using 32-dimensional hidden layer is sufficient for dynamic adapters in DASD.
%We can see that the performance is optimal when the $d_u$ is 32. Reducing the $d_u$ leads to a significant drop in performance, while increasing the $d_u$ beyond 32 results in only marginal performance improvements. 
\input{table_ablation_hiddensize}

\begin{figure}[tb!]
\centering
\includegraphics[width=0.48\textwidth]{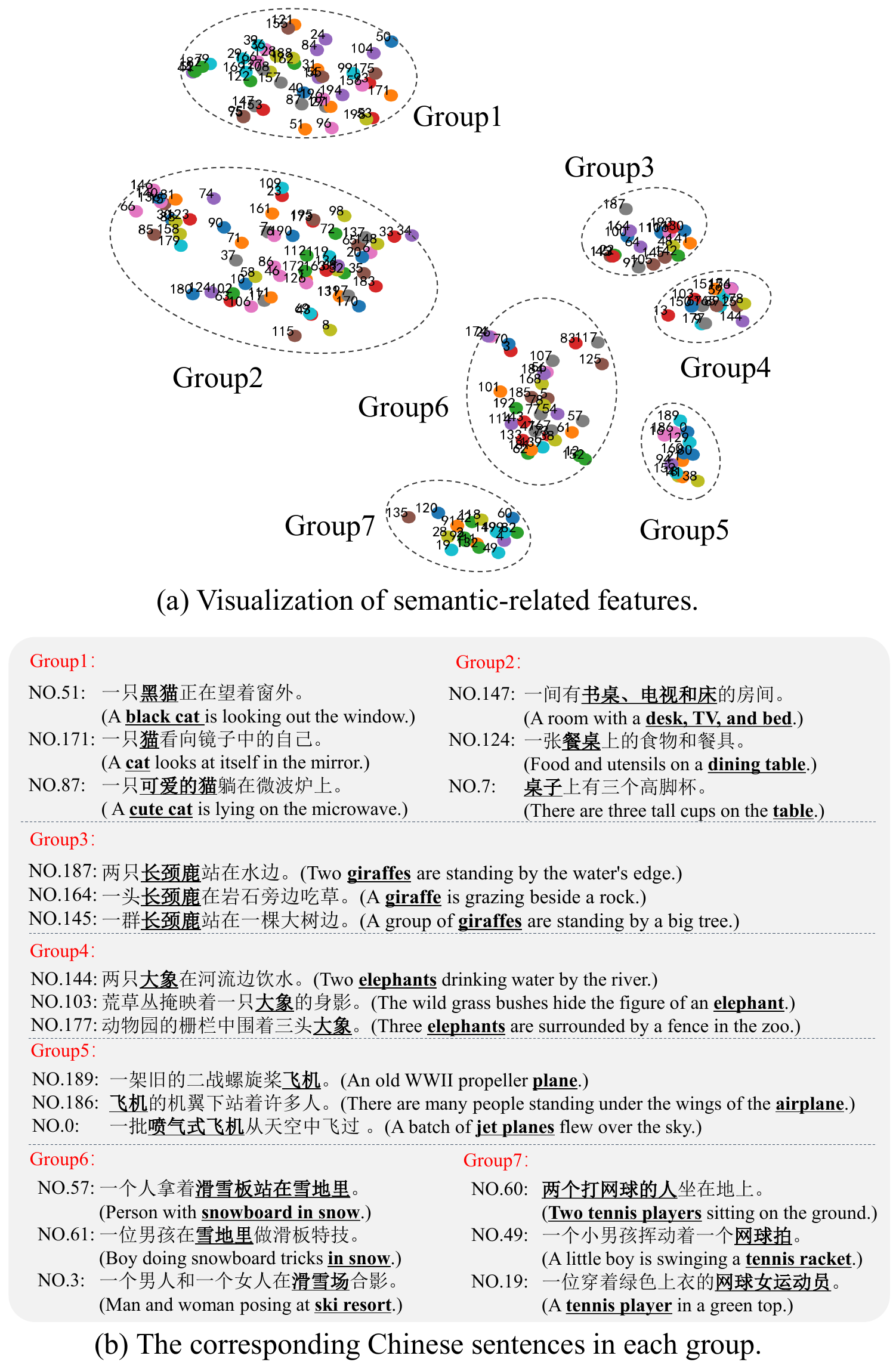}
\caption{Visualization of the semantic-related features extracted from 200 randomly-selected Chinese sentences in MSCOCO testset.}\label{fig:tsne_sr}
\end{figure}

\begin{figure*}[]
\centering
\includegraphics[width=0.9\textwidth]{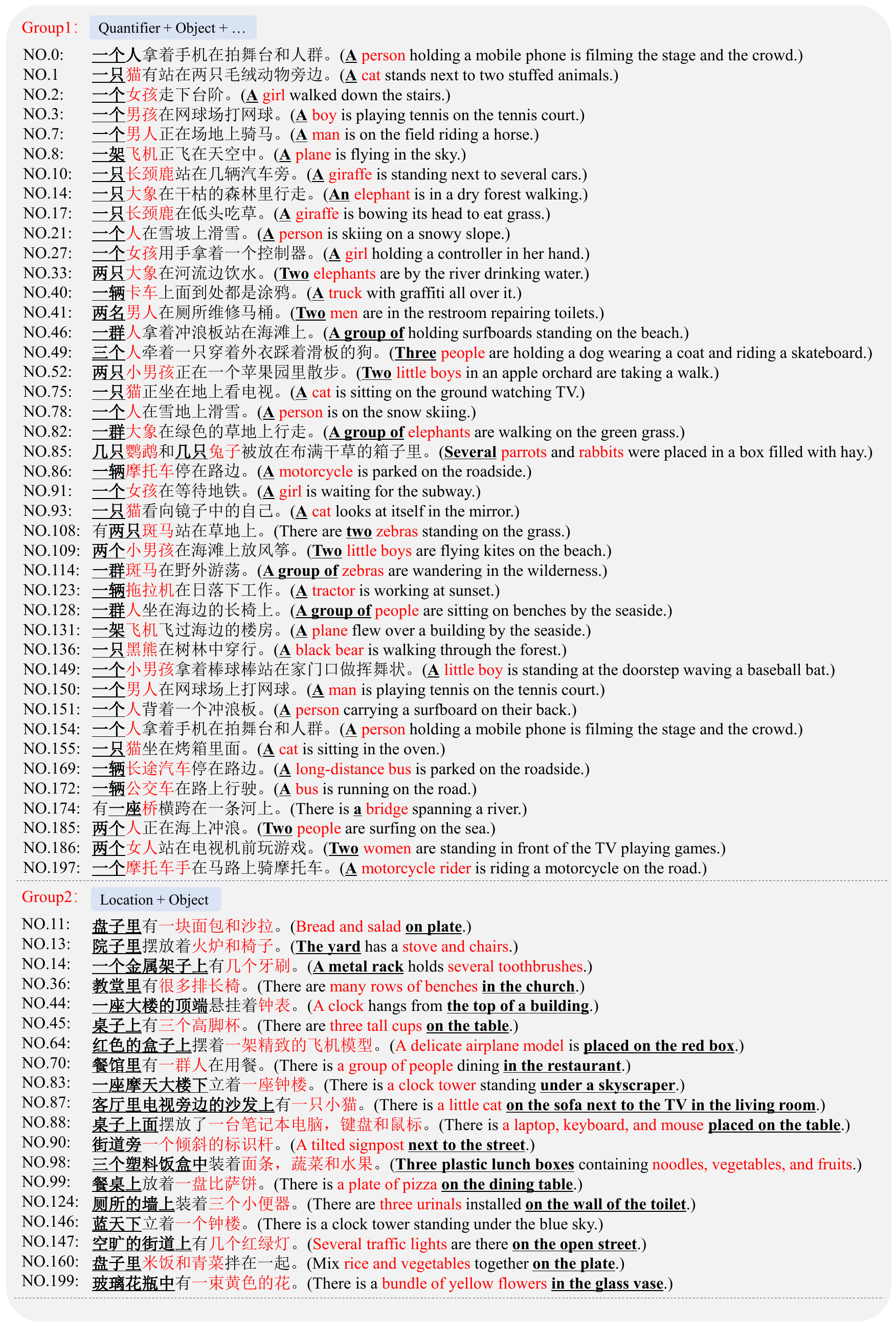}
\caption{Supplementary samples within the scope of group 1 and group 2 in the main text Figure 3(a).}
\label{fig:text1}
\end{figure*}

\begin{figure*}[]
\centering
\includegraphics[width=0.92\textwidth]{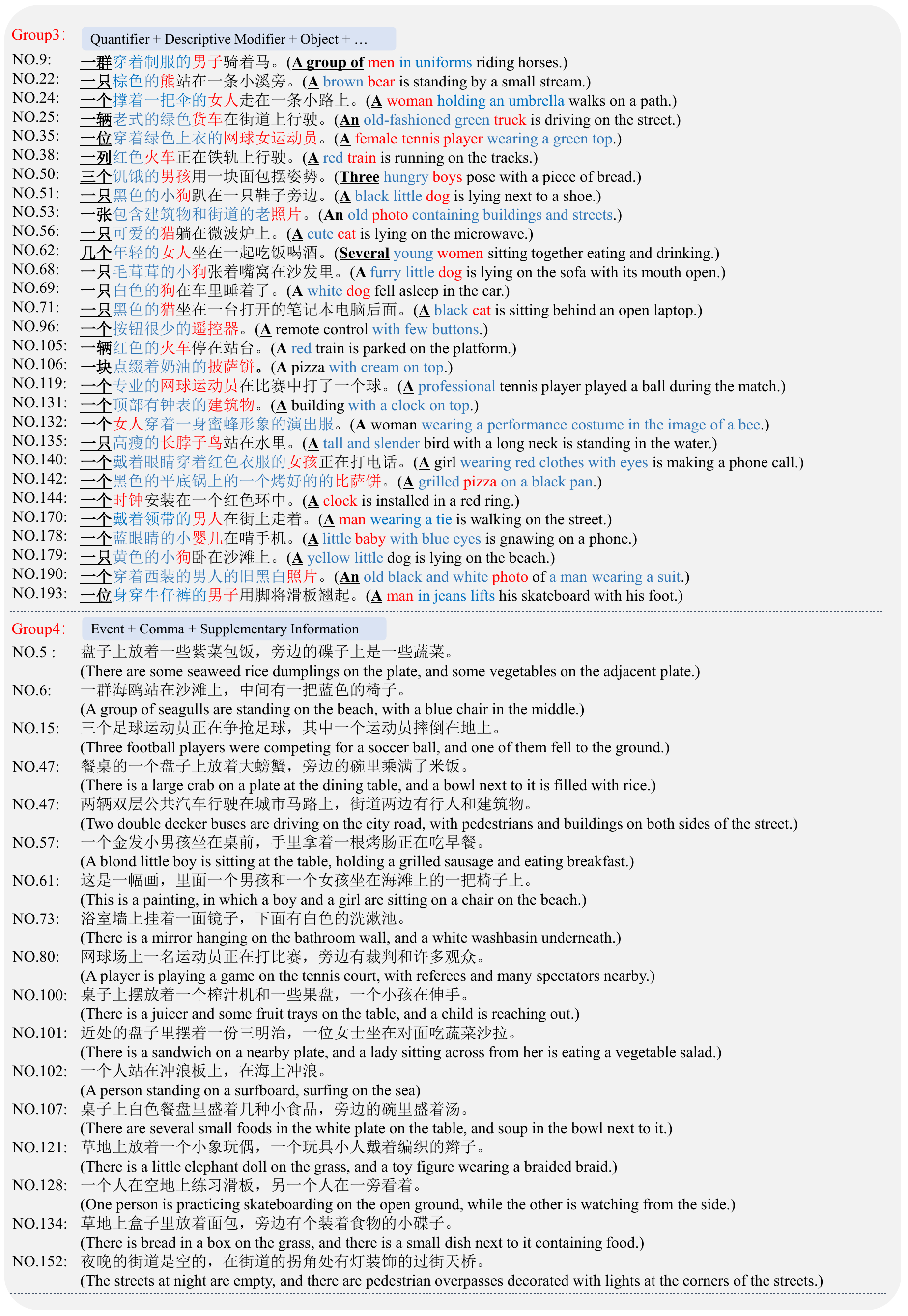}
\caption{Supplementary  samples within the scope of group 3 and group 4 in the main text Figure 3(a).}
\label{fig:text2}
\end{figure*}

\subsection{Visualization of Representations}\label{sec:visual}
\subsubsection{Semantic-related Features Visualization}
Similar to the visualization of the semantic-agnostic features in the main text, we also perform a t-SNE visualization of the semantic-related features. 
As shown in Figure~\ref{fig:tsne_sr}(a), 
features of 200 sentences are clearly clustered into seven groups, and  some corresponding sentences of each group are listed in Figure~\ref{fig:tsne_sr}(b).
It is obvious that sentences clustered into the same group share a common semantic topic.
For example,  group 1 contains sentences describing kittens, while sentences in group 2 are all about furniture such as tables and chairs.
The result of visualization demonstrates that the semantic-related features learned by DASD are highly related to the semantics of their corresponding captions, verifying the effectiveness of semantic consistency learning.
%Sentences in Group 3 and Group 4  are about giraffes and elephants, respectively.
%These two groups are closer than others maybe due to the presence of large animals in their sentences.
%Group 5 consists of sentences describing airplanes or airports. Finally, Group 6 and Group 7 contain sentences about tennis and skiing, respectively.
%Sentences in each group are compacted clustered, and  the semantic-related features have been explicitly aligned with the semantic features produced by CLIP through semantic consistency learning,  the clustering results are more pronounced compared to the semantic-agnostic features.

% \begin{figure*}[tb!]
% \centering
% \includegraphics[width=0.88\textwidth]{fig/tsne_text1.pdf}
% \caption{Sentences within the scope of Group 1 and Group 2 shown in the main text Figure 3(a).}
% \label{fig:text1}
% \end{figure*}

% \begin{figure*}[tb!]
% \centering
% \includegraphics[width=0.88\textwidth]{fig/tsne_text2.pdf}
% \caption{Sentences within the scope of Group 3 and Group 4 shown in the main text Figure 3(a).}
% \label{fig:text2}
% \end{figure*}

\subsubsection{Semantic-agnostic Features Visualization}
The details of the semantic-agnostic feature visualization have  been introduced in the main text. 
Due to space limitations, only three samples in each of the four groups are displayed in the main text Figure 3(b). 
Here we supplement more samples in these groups for readers, shown in Figure~\ref{fig:text1} and Figure~\ref{fig:text2}.
We observe that most sentences conform to the clustering characteristics of each groups described in the main text, further verifying the effectiveness of adversarial training in semantic disentangling.

%% file: table_ablation_LossFunction.tex
\begin{table}[tb!]
\centering
\renewcommand{\arraystretch}{1.2}
\caption{Performance of our model using different semantic consistency loss function.}
\vspace{-3mm}

\small
\begin{tabular}{lccccccc}
\hline
\multirow{2}{*}{\textbf{loss}}   & \multicolumn{3}{c}{\textbf{Multi30K}}         &           & \multicolumn{2}{c}{\textbf{MSCOCO}} & \multirow{2}{*}{\textbf{SUM}} \\ \cline{2-4} \cline{6-7}
                 & \textbf{FR}            & \textbf{DE}            & \textbf{CS}            &           & \textbf{ZH}             & \textbf{JA}             &                               \\ \hline
L1(ours)               & 91.1          & 88.5          & 87.6          &           & 90.0           & 89.1           & 446.3                         \\
L2               & 89.5 & 88.2 & 87.1 & \textbf{} & 89.4  & 88.8  & 443.0                \\
Smooth L1        & 91.0          & 88.2          & 87.3          &           & 89.6           & 88.7           & 444.8                         \\ \hline
\end{tabular}
\label{con_loss}
\end{table}

%% file: table_ablation_CC3M.tex
\begin{table}[tb!]
\centering
\renewcommand{\arraystretch}{1.2}
\caption{Model performance using training datasets of different sizes.}
\vspace{-3mm}
\small
\begin{tabular}{lccccccc}
\hline
\multirow{2}{*}{\textbf{Datasets}}   & \multicolumn{3}{c}{\textbf{Multi30K}}         &           & \multicolumn{2}{c}{\textbf{MSCOCO}} & \multirow{2}{*}{\textbf{SUM}} \\ \cline{2-4} \cline{6-7}
                 & \textbf{FR}            & \textbf{DE}            & \textbf{CS}            &           & \textbf{ZH}             & \textbf{JA}             &                               \\ \hline
CC300K           & 88.6          & 87.4          & 83.4          &           & 88.5           & 84.8           & 432.7                             \\
CC3M             & 88.7          & 87.6          & 84.3          &           & 88.7           & 85.3           & 434.6                            \\
 \hline
\end{tabular}
\label{table_3M}
\end{table}

%% file: table_ablation_drop_ClipLayer.tex
\begin{table}[tb!]
\centering
\renewcommand{\arraystretch}{1.2}
\caption{Model performance with different numbers of clip layers in the target-language branch.}
\vspace{-3mm}

\small
\begin{tabular}{cccccccc}
\hline
\multirow{2}{*}{\textbf{\#Clip Layers}}   & \multicolumn{3}{c}{\textbf{Multi30K}}         &           & \multicolumn{2}{c}{\textbf{MSCOCO}} & \multirow{2}{*}{\textbf{SUM}} \\ \cline{2-4} \cline{6-7}
                 & \textbf{FR}            & \textbf{DE}            & \textbf{CS}            &           & \textbf{ZH}             & \textbf{JA}             &                               \\ \hline
1        & 59.3          & 61.4          & 50.3          &           & 61.7           & 59.7           & 292.4                          \\
3        & 64.0          & 66.1          & 64.4          &           & 72.4           & 66.9           & 333.8                          \\
6        & 70.2          & 72.4          & 70.1          &           & 81.9           & 72.5           & 367.1                          \\
9        & 81.7          & 82.7          & 83.1          &           & 84.0           & 80.1           & 411.6                          \\
11       & 87.5          & 87.9          & 86.4          &           & 89.8           & 88.3           & 439.9                          \\
12       & 91.1          & 88.5          & 87.6          &           & 90.0           & 89.1           & 446.3                         \\
 \hline
\end{tabular}
\label{clip_layers}
\end{table}

%% file: table_ablation_woDA.tex
\begin{table}[tb!]
\centering
\renewcommand{\arraystretch}{1.2}
\caption{Model performance without dynamic adapters distributed at a certain CLIP layer.}
\vspace{-3mm}
\small
\begin{tabular}{cccccccc}
\hline
\multirow{2}{*}{\textbf{\#Clip Layer}}   & \multicolumn{3}{c}{\textbf{Multi30K}}         &           & \multicolumn{2}{c}{\textbf{MSCOCO}} & \multirow{2}{*}{\textbf{SUM}} \\ \cline{2-4} \cline{6-7}
                 & \textbf{FR}            & \textbf{DE}            & \textbf{CS}            &           & \textbf{ZH}             & \textbf{JA}             &                               \\ \hline
-           & 91.1          & 88.5          & 87.6          &           & 90.0           & 89.1           & 446.8                            \\
12th        & 90.4          & 87.3          & 86.8          &           & 89.4           & 88.6           & 442.5                           \\
11th        & 90.7          & 88.1          & 87.4          &           & 89.9           & 88.9           & 445.0                          \\

10th        & 90.9          & 88.5          & 87.6          &           & 90.0           & 89.0           & 446.0                          \\

9th         & 91.1           & 88.3          & 86.9          &           & 90.0           & 88.6           & 444.9                          \\
8th         & 90.8          & 88.5          & 87.5          &           & 89.7           & 88.7           & 445.2                          \\
7th         & 90.7          & 88.3          & 87.2          &           & 89.9           & 88.9           & 445.0                          \\
6th         & 91.0          & 87.6          & 87.4          &           & 89.5           & 88.4           & 443.9                          \\
5th         & 90.9          & 87.7          & 87.4          &           & 88.9           & 88.9           & 443.8                          \\
4th         & 91.0          & 87.6          & 87.3          &           & 89.7           & 89.1           & 444.7                          \\
3rd         & 90.6          & 88.2          & 87.2          &           & 89.9           & 89.0           & 444.4                          \\
2nd         & 91.1          & 87.8          & 87.4          &           & 89.9           & 88.7           & 444.9                          \\
1st         & 90.9          & 87.7          & 87.3          &           & 89.6           & 88.8           & 444.3                          \\
\hline
\end{tabular}
\label{table_DA}
\end{table}

%% file: table_ablation_layers.tex
\begin{figure}[t]
\hspace*{-.2cm}\begin{tikzpicture}[scale=.80]
\pgfplotsset{every axis/.append style={line width=1pt}}

\begin{axis}[
height=6cm,
width=10cm,
  ymax=89.5,
  ymin=88.3,
  xmax=5,
  xmin = 0,
ylabel={mAR}, xlabel={Number of pretrained layers in SDM}, xtick={0,1,...,5},xticklabels = {0,
             1,          
             3,          
             6,         
             9,
            12},x tick label style={rotate=0,anchor=north}]
    \addplot[color=blue,mark=*] coordinates {
       (0,88.8)
      (1,89.3)  (2,89.1) 
       (3,89.2)  (4,89.1) (5,89.1)};
    \addplot[color=gray,style=dashed, mark=] coordinates {
       (0,89.0)
       (5,89.0)};    

\end{axis}

\end{tikzpicture}
\caption{Performance of our model  varies with the number of pretrained layers employed for semantic disentangling.}
\label{fig:Klayer}
\end{figure}

%% file: table_ablation_hiddensize.tex
\begin{table}[tb!]
\centering
\renewcommand{\arraystretch}{1.2}
\caption{Performance of our model using dynamic adapters with various hidden-layer dimensions.}
\vspace{-3mm}

\small
\begin{tabular}{cccccccc}
\hline
\multirow{2}{*}{\textbf{Dimension}}   & \multicolumn{3}{c}{\textbf{Multi30K}}         &           & \multicolumn{2}{c}{\textbf{MSCOCO}} & \multirow{2}{*}{\textbf{SUM}} \\ \cline{2-4} \cline{6-7}
                 & \textbf{FR}            & \textbf{DE}            & \textbf{CS}            &           & \textbf{ZH}             & \textbf{JA}             &                               \\ \hline
16               & 90.2          & 87.9          & 87.1          &           & 89.6           & 88.2           & 443.0                         \\
32               & 91.1 & 88.5 & 87.6 & \textbf{} & 90.0  & 89.1  & 446.3                \\
64               & 91.4          & 88.2          & 87.5          &           & 90.2           & 89.3           & 446.6                         \\ \hline
\end{tabular}
\label{tab:hidden_size}
\end{table}